\documentclass{article}

 \usepackage[preprint]{neurips_2025}

\usepackage{times}
\usepackage{latexsym}
\usepackage{microtype}
\usepackage{inconsolata}
\usepackage{float}
\usepackage{graphicx}
\usepackage{amsmath}
\usepackage{amssymb}
\usepackage[linesnumbered,ruled,vlined]{algorithm2e}
\usepackage{caption}
\usepackage{subcaption}
\usepackage{multirow}
\usepackage{enumitem}
\usepackage{xspace}
\usepackage{wrapfig}
\usepackage{tablefootnote}
\usepackage{fontawesome5}
\pdfmapline{+markerfelt < Markerfelt.ttc <T1-WGL4.enc}

\newcommand\MoviePuzzleTitle{MoviePuzzle \xspace}
\newcommand\MoviePuzzle{MoviePuzzle \xspace}

\makeatletter
\renewcommand{\paragraph}{%
  \@startsection{paragraph}{4}%
  {\z@}{0ex \@plus 0ex \@minus 0ex}{-1em}%
  {\hskip\parindent\normalfont\normalsize\bfseries}%
}
\makeatother

\makeatletter
\DeclareRobustCommand\onedot{\futurelet\@let@token\@onedot}
\def\@onedot{\ifx\@let@token.\else.\null\fi\xspace}

\def\ie{\textit{i.e}\onedot} 
 
\def\etc{\textit{etc}\onedot} 
\def\wrt{w.r.t\onedot} 

\makeatother

\usepackage[utf8]{inputenc} % allow utf-8 input
\usepackage[T1]{fontenc}    % use 8-bit T1 fonts
\usepackage[pagebackref,breaklinks,colorlinks,allcolors=cvprblue]{hyperref}
\usepackage{url}            % simple URL typesetting
\usepackage{booktabs}       % professional-quality tables
\usepackage{amsfonts}       % blackboard math symbols
\usepackage{nicefrac}       % compact symbols for 1/2, etc.
\usepackage{microtype}      % microtypography

\usepackage{orcidlink}

\usepackage[capitalize]{cleveref}
\usepackage[markup=underlined,defaultcolor=red,addedmarkup=uwave]{changes}

\Crefname{equation}{Eqn.}{Eqns.}
\Crefname{section}{Sec.}{Secs.}
\Crefname{figure}{Fig.}{Figs.}
\Crefname{table}{Tab.}{Tabs.}
\crefname{table}{Tab.}{Tabs.}

\newcommand\HCMC{{\textsc{HCMC}}\xspace}
\newcommand\Clip{{\textsc{Clip}}\xspace}
\newcommand\Bert{{\textsc{Bert}}\xspace}
\newcommand\AlBert{{\textsc{AlBert}}\xspace}
\newcommand\VideoMAE{{\textsc{VideoMAE}}\xspace}
\newcommand\Timesformer{{\textsc{Timesformer}}\xspace}
\newcommand\Singularity{{\textsc{Singularity}}\xspace}
\newcommand\VideoLLaVA{{\textsc{VideoLLaVA}}\xspace}
\newcommand\GPT{{\textsc{GPT}}\xspace}
\newcommand\ChatGPT{{\textsc{ChatGPT}}\xspace}
\newcommand\LLavaNext{{\textsc{LLavaNext}}\xspace}
\newcommand\TimeChat{{\textsc{TimeChat}}\xspace}
\newcommand\Vtimellm{{\textsc{Vtimellm}}\xspace}

\newcommand{\review}[1]{{}}

\definecolor{cvprblue}{rgb}{0.21,0.49,0.74}
\title{\texorpdfstring{\faPuzzlePiece}{}\MoviePuzzleTitle: Visual Narrative Reasoning through Multimodal Order Learning} 

\author{%
  Jianghui Wang$^{1\ast}$, Yuxuan Wang$^{2\ast}$, Dongyan Zhao$^{2}$, Zilong Zheng$^{1\dagger}$ 
  \thanks{Equal contribution. $\dagger$ Corresponding author.}\vspace{5pt}\\
$^1$ Beijing Institute for General Artificial Intelligence, Beijing, China \\
$^2$ Wangxuan Institute of Computer Technology, Peking University, Beijing, China \vspace{5pt}\\
\url{https://moviepuzzle.github.io}
  \vspace{-25pt}
}

\begin{document}
\maketitle
\begin{abstract}

In recent years, video-language understanding has seen rapid progress. However, existing studies often overlook the need for \textbf{holistic video understanding} and the \textbf{inherent narrative structure} within long-form videos. To bridge this gap, we present MoviePuzzle, a novel benchmark that promotes visual narrative reasoning (VNR) by reshuffling the shot, frame, and clip layers of movie segments, guided by video-dialogue context. We construct a carefully curated dataset where movies are decomposed into hierarchical narrative layers, with their temporal order randomly shuffled. In addition to benchmarking MoviePuzzle with existing approaches, we propose Hierarchical Contrastive Movie Clustering (HCMC), a model that takes advantage of structural and semantic cues to recover narrative coherence. HCMC employs pairwise and contrastive learning to model narrative flow and resolve temporal disorder in videos. We assess performance through both upstream and downstream tasks, and introduce a sub-sequence-based ordering score to evaluate temporal alignment. Experimental results show that our approach outperforms state-of-the-art baselines on the MoviePuzzle benchmark, underscoring its effectiveness in capturing visual narrative structure. Further study of our method on general video understanding tasks demonstrates its potential to enhance the performance of existing approaches, extending its capability to improve generative models. 

\end{abstract}

\section{Introduction}\label{sec:intro}

\begin{figure*}[t!]

\includegraphics[width=\linewidth]{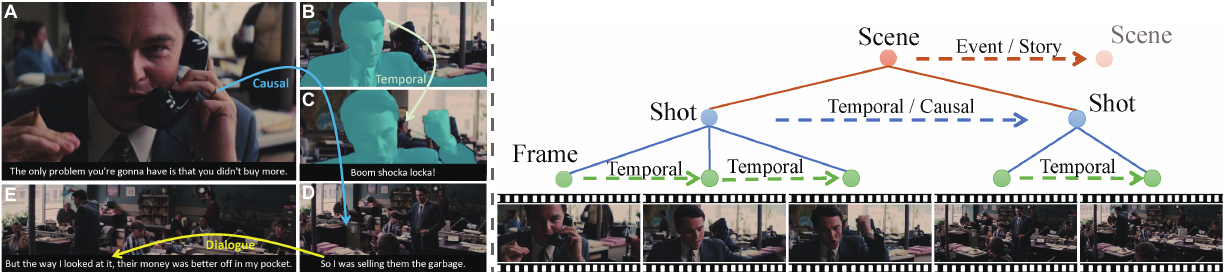}
    \caption{
    \textbf{Illustration of MoviePuzzle task and visual narrative structures.} Left: An example of our benchmark, where arrows highlight temporal, causal and dialogue clues across frames. Right: The 3-layer visual narrative structure of the clip; refer to $\S$\ref{sec:moive_hierachy} for hierarchical representation. 
    }
    \label{fig:structure}
    % \vspace{-.15in}
\end{figure*}

Humans—even young children—can effortlessly perceive and interpret various forms of visual media, including comics, short videos, and 3D films. Without attending to fine-grained details, we intuitively connect salient visual and auditory signals to construct a coherent \textit{visual narrative}~\cite{cohn2013visual} in real time.
Consider the movie frames in~\Cref{fig:structure}; one can easily infer the storyline—\textit{the man is scamming money over the phone}—without needing to count the exact number of individuals in panel \textbf{E}. This cognitive ability enables us to understand long-form content such as movies and entire seasons of television shows.
Our work centers on preserving narrative coherence, rather than simply reordering shots for stylistic non-linear storytelling, as is common in montage-based editing. We argue that non-linear elements~\cite{frey2021non, park2018non} should ultimately reinforce the overarching narrative by maintaining continuity and logical progression. To further evaluate the quality of our model’s outputs, we conduct comprehensive human studies assessing the plausibility of the generated sequences.

In the computer vision community, video understanding has witnessed significant improvements in recent years, supported by various benchmarks proposed on fine-grained action predictions~\cite{yeung2016end,feichtenhofer2020x3d}, video question answering~(VideoQA)~\cite{fu2021violet,lei2019tvqa,movieqa}, and video-grounded dialogue generations~\cite{le2020video,le-etal-2022-vgnmn,wang2023vstar}, \etc. The \textit{de facto} paradigm is to integrate all extracted dense features of vision and language into a fusion layer \wrt their temporal orders~\cite{xu2015show,anderson2018bottom}. Despite promising results from video-language models, current benchmarks often overlook three critical dimensions: 

\textit{Video Dialogue Temporal Learning.} In NLP, The importance of sequence ordering is well-established. Pretraining framework such as ALBERT and ERNIE~\cite{albert, ernie-doc} emphasize temporal coherence, while dialogue systems rely on utterance sequencing for effective interaction~\cite{DBLP:conf/sigdial/Lison11, DBLP:conf/eacl/GalitskyI17, DBLP:conf/ijcai/KhouzaimiLL16,DBLP:conf/aaai/QuanY0XLDOTDLYJ21,DBLP:journals/ijsr/GrassiRS22}. Reordering benchmarks like~\citep{Li2023LooGLECL,li2024temporal}further highlight long-context modeling as a key capability. In CV, visual reordering tasks are recognized as beneficial for video comprehension ~\cite{sharma2020deep,epstein2021learning,yeung2016end}, yet temporal learning remains underexplored in multimodal dialogue tasks. This gap undermines the integration of shared temporal dynamics between dialogue and video streams.

\textit{Holistic Video Dialogue Understanding.} Most existing work abstracts video understanding into multi-frame image modeling, focusing on frame-level and short-span interactions. This reductionist view limits the scalability to long-form content, where scenes and shot transitions are central. Although recent efforts such as MovieNet~\cite{huang2020movienet} have emphasized holistic understanding, the tasks are still framed as recognition problems. Similarly, holistic dialogue understanding is gaining traction~\cite{Song2016DialogueSS, Xu2021TopicAwareMD,Xing2021ImprovingUD}, particularly in modeling topic transitions. However, few studies attempt to unify these perspectives for comprehensive video-dialogue understanding.

\textit{Visual Narrative Structures.} Beyond individual visual cues or subtitle tokens, the interpretation of video sequences hinges on underlying\textit{ narrative structures}—hierarchical frameworks that govern the unfolding of events and concepts~\cite{cohn2013visualbook}. As illustrated in~\Cref{fig:structure}, each frame contributes to the overall story; disrupting their order may compromise narrative integrity. Interestingly, alternative reorderings can yield plausible yet distinct narratives, suggesting broad applicability for video reordering in tasks such as automated editing, video synthesis, and more general AGI challenges.

To address the limitations of prior work and advance holistic understanding of long-form videos, we introduce MoviePuzzle, a new benchmark designed to evaluate machine capabilities in visual narrative reasoning (VNR). In this task, models are challenged to reorder frame sequences into coherent storylines—see~\Cref{fig:structure} for an example. Our dataset is built upon MovieNet~\cite{huang2020movienet}, with precise alignment between video frames and subtitles to ensure narrative clarity. To keep the task computationally tractable, we constrain the number of scene and shot transitions per clip. The choice of frame reordering as the core task is motivated by diagnostic insights into the role of narrative structure in visual storytelling~\cite{cohn2013visualbook}. Experiments show that using our task as a pretraining step improves performance on multiple downstream video understanding benchmarks. This highlights the benchmark’s value in enhancing temporal reasoning and its potential for guiding MLLM design.visual narratives.

\section{The MoviePuzzleTitle Dataset}\label{sec:dataset}

The MoviePuzzle contains a diverse range of data from various modalities and rich annotations on different aspects of movie structures, enabling a comprehensive understanding of movie content. Below we introduce detailed data curation and annotation.
\begin{figure*}[t!]
    \centering
\begin{minipage}[t!]{.335\linewidth}
    \centering
    \includegraphics[width=\linewidth]{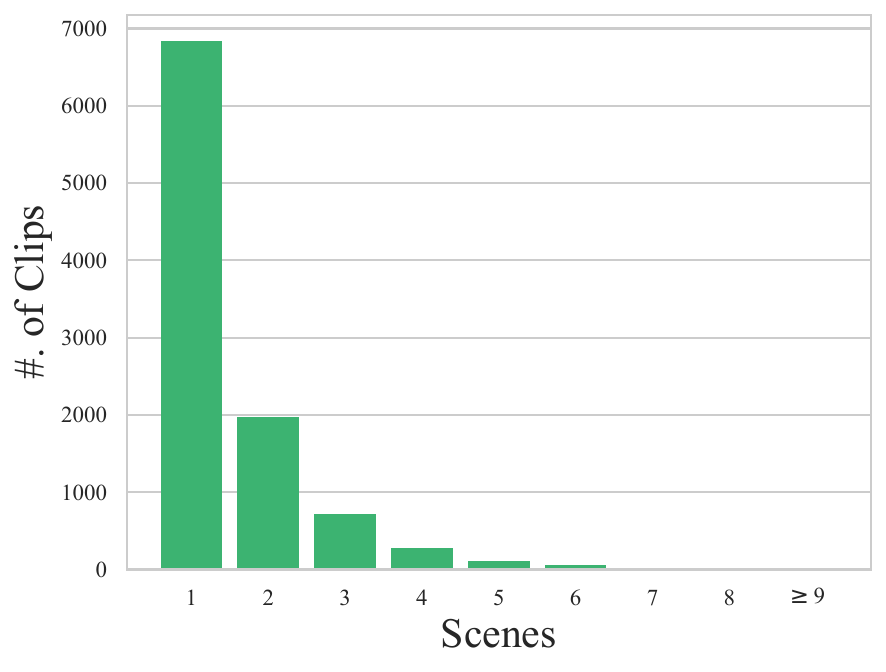}
    % \vspace*{-0mm}
    \captionof{figure}{Scenes Distribution}
    % {Distribution of scenes in a clip.}
    \label{fig:statistic_clip-scene}
\end{minipage}
\hfill
\begin{minipage}[t!]{.335\linewidth}
    \centering
    \includegraphics[width=\linewidth]{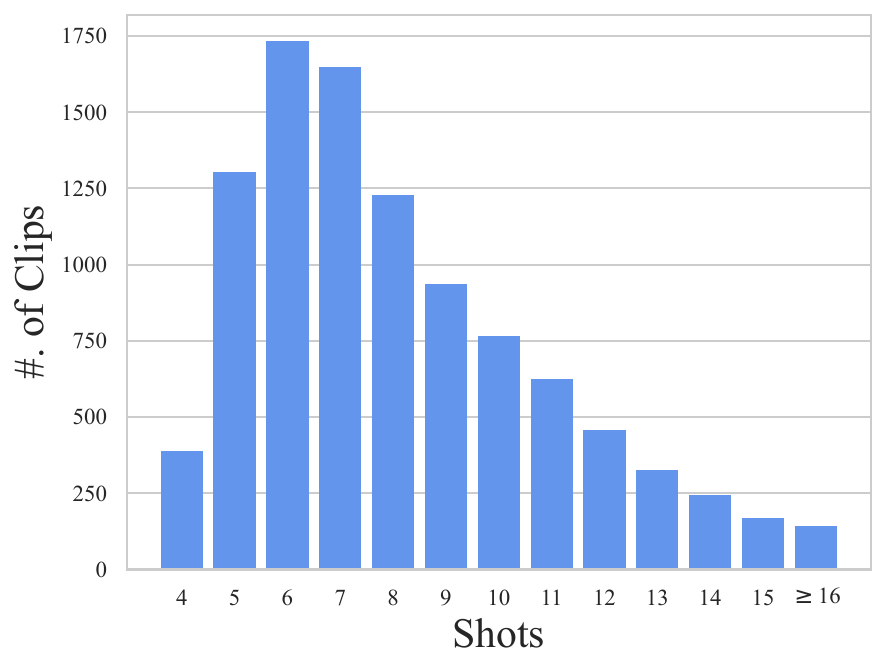}
    % \vspace*{-0mm}
    \captionof{figure}{Shots distribution}
    % {Distribution of shots in a clip. } 
    \label{fig:statistic_clip-shot}
\end{minipage}
\hfill
\begin{minipage}[t!]{.31\linewidth}
    \centering
    \includegraphics[width=\linewidth]{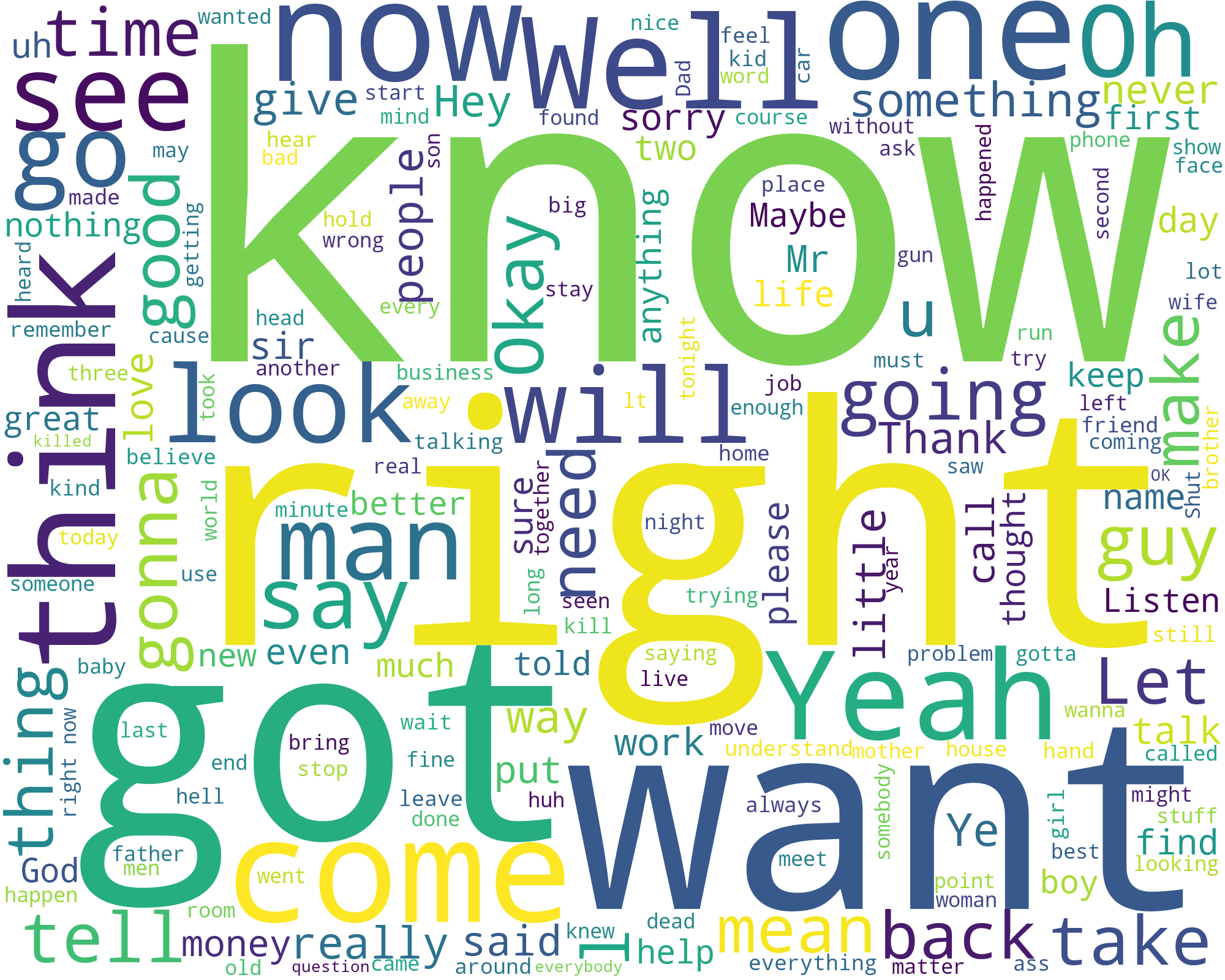}
    % \vspace*{-0mm}
    \captionof{figure}{Wordcloud}
    % {Wordcloud of utterances.} 
    \label{fig:wordle}
\end{minipage}
 
\end{figure*}

\subsection{Dataset Source}
Utilizing MovieNet \cite{huang2020movienet}, we curated a subset of 228 films with comprehensive multimodal annotations including frame images, subtitles, and structural boundaries for scenes and shots, alongside synopses. From this subset, we extracted 135,837 frames as RGB images, cropped to eliminate black borders, from 240P quality films. English subtitles, filtered for synchronization with the movies and purged of non-Unicode symbols and foreign characters outside the standard English alphabet, were also collected, excluding movies with fewer than 500 dialogues and subtitles with non-verbal sounds. The resulting data presents challenges in aligning and understanding dialogue with visuals, as depicted in the word cloud of training utterances in \Cref{fig:wordle}.

\subsection{Curated Data Annotation}

\paragraph{Image-utterance pairs}
We pair each frame image with its corresponding utterance, achieving the finest granularity alignment of images and text on the time sequence. Each dialogue is only matched with the images falling exactly within its corresponding time period, rather than the nearest ones, ensuring the maximum matching of image and text content. In the case of multiple frames falling on a single dialogue, we choose to merge these frames and select the middle frame as the most representative matching image. For most original images that do not match subtitles, we discard them unless they are exceptional examples within a clip, filtering out 16.5\% of the data.

\paragraph{Scene\&Shot aligned boundary}
Regarding the visual narrative structure of movies as in \Cref{fig:structure}, most clips have two innate hierarchical levels: shots and scenes. 
A shot is a fixed camera angle capture and typically a short sequence of temporally consecutive frames indicating a few actions. 
A scene is a sequence of shots that share the same environmental context, which typically depicts an event or a short story.
Capturing the hierarchical structure of a movie is vital for movie understanding. 
Of note, scene segmentation remains an open problem to video understanding~\cite{chen2022movies2scenes,rasheed2003scene,rao2020local,chen2021shot}. We leverage annotations from MovieNet~\cite{huang2020movienet} to form our labels, including the manually annotated scene boundaries and automatically generated shot boundaries using \cite{shao2018find}. In total, the MoviePuzzle has 15K scenes and 81K semantically shots.

\paragraph{Clip aligned boundary}
In order to obtain video clips suitable for training, we further segment the 228 movies and ultimately obtain 10,031 movie clips. We use the following two criteria to select appropriate clips as our data: First, each clip should consists of 10 to 20 frames, and the length of the clip is controlled to tell a short plot without the semantic information being too monotonous; Second, the image-text pairs in each clip must be greater than 80\%, ensuring the semantics are not too sparse. As shown in \Cref{fig:statistic_clip-scene,fig:statistic_clip-shot},
most clips have 1 or 2 scenes (88\%) and 5 to 10 separated shots (75.91\%), aligning with a typical short plot structure.

\subsection{Data Statistics}
We divide the entire MoviePuzzle dataset into four splits: train, val, in-domain test, and out-domain test, resulting in 7,048/589/1,178/1,196 clips, respectively. 
% each accounting for 70\%, 6\%, 12\%, and 12\% of the total clips, respectively. \zzl{ \#. of Clips } 
% 10,031
The data in val and in-domain test come from the same set of movies as the train split. Specifically, we select val and in-domain test by taking equally spaced clips based on the clip numbers from the movie clips covered by the train split. The out-domain test split, sourced from different movies than the train split, assesses model generalization.

% Methodology
\section{Benchmarking the MoviePuzzleTitle}

\subsection{Task Formulation}\label{sec:task-form}

Formally, each clip $C$ within the MoviePuzzle dataset $\mathcal{D}$ can be denoted as a sequence of randomly shuffled but temporally aligned vision-utterance pairs, \ie, $C = \{(v_1, u_1), (v_2, u_2), \cdots, (v_{N_{fm}}, u_{N_{fm}})\} \in \mathcal{D}$, with $v_i$ and $u_i$ index the $i$-th frame and utterance, $N_{fm}$ denotes the number of total frames within a clip. Besides, each frame is labeled with the corresponding shot id $st_i$ and scene id $sn_i$, identifying the unique shot and scene classes within the movie clip. The goal of the MoviePuzzle challenge is to predict each frame's temporal index $\{l_i\}_{i=1}^{N_{fm}}$. Especially, for a temporally ordered sequence, $idx_i = i$ for all frames. 

\paragraph{Hierarchical Movie Representations}\label{sec:moive_hierachy}

In this work, we represent an ordered movie clip as a compositional visual narrative structure. From bottom to top, a clip can be represented within frame-level, shot-level, and scene-level; \Cref{fig:structure} depicts the overall representation. 

\begin{enumerate}
    \item \textit{Frame-Level Representation} We employ $F = \{f_1, f_2, \cdots, f_{N_{fm}}\}$ to denote the frame-level representation of a clip $C$, where any $f_k \in F$ encompasses all information contained within the $k$-th frame. It includes the visual data $v_k$, textual information $u_k$, and the corresponding shot identity $st_k$ to which the frame belongs.

    \item \textit{Shot-Level Representation} For the shot-level representation of a clip, we utilize $G = \{g_1, g_2, \cdots, g_{N_{st}}\}$ to represent, where any $g_j \in G$ incorporates all information present in the $j$-th shot, and $N_{st}$ signifies the quantity of shots employed in the referred movie clip. This comprises a set of frames $\{f\}_{k=1}^{N_{st}^j}$, with ${N_{st}^j}$ indicates the frames number in the $j$-th shot. Each represents a frame $f_k \in F$ within the shot $g_j$, and its associated scene identity $sn_j$.

    \item \textit{Scene-Level Representation} Similarly, the scene-level representation of a clip is characterized by $H=\{h_1, h_2, \cdots, h_{N_{sc}}\}$, with any $h_i \in H$ containing all information within the $i$-th scene. This involves a set of shots $\{g\}_{j=1}^{N_{sn}^i}$ denoting each shot $g_j \in G$ in the scene, where ${N_{sn}^i}$ indicates the shot number in scene $g_j$.

\end{enumerate}

\paragraph{Evaluation Metrics}\label{sec:metric}
We consider four classical metrics on temporal reordering: Location Square Deviation (LSD) and Location Mean Deviation (LMD) compute the average value of square and absolute differences in positions of two sequences; Swap Deviation (SD) and Swap Distance Deviation (SSD) compute the minimum number of swaps and transformation weight to convert from predicted sequence to ground truth. All metrics indicate better performance with lower values.

Moreover, to mitigate bias and address the absence of ordering from generative models, which greatly constrains the usage of existing metrics, we propose a new, robust sub-sequence-based evaluation metric, Ordering Score (OS), which is computed with the number of matched sub-sequences for each pair ($\beta=2$) or each triple ($\beta=3$). Higher OS indicates a greater ratio of matched sub-sequences. Therefore, we propose a new metric for temporal sequence evaluation.

% As defined in \Cref{sec:task-form}, 
The video clip has $N_{fm}$ frames, the ground truth sequence is $\{ l_1, l_2, \cdots, l_{N_{fm}}\}$, and model ordering prediction sequence is $\{\hat{l}_1, \hat{l}_2, \cdots, \hat{l}_{N_{fm}}\}$. We define a match as a $\beta$ length subset of elements that follow their natural order: $i_1 < i_2 < \cdots < i_\beta$ and $l_{i_1} \prec l_{i_2} \prec \cdots \prec l_{i_\beta}$ and $\hat{l}_{i_1} \prec \hat{l}_{i_2} \prec \cdots \prec \hat{l}_{i_\beta}$, where $\prec$ is a comparator indicating the precedent temporal order. The ordering score is computed as the ratio of the cardinality of the match set:

\begin{equation}
    \text{OrderingScore} = \frac{|\mathcal{M}|}{{\tbinom {N_{fm}} \beta}}
\end{equation}
where $\mathcal{M} = \{(i_k)_{k=1}^{n} | (i_1 < \cdots < i_\beta) \wedge (l_{i_1} \prec \cdots \prec l_{i_\beta}) \wedge (\hat{l}_{i_1}  \prec \cdots \prec \hat{l}_{i_\beta}) \}$ represents the index set of all sequences that satisfy the order matching., and ${\tbinom {N_{fm}} \beta} = \frac{N_{fm}!}{\beta!(N_{fm}-\beta)!}$ is the total number of index combinations.

\paragraph{Pairwise Score} The pairwise matching score is one of the most commonly used metrics for permutations in discrete mathematics and permutation graph theory.

\paragraph{Triplet Score} The pairwise score can be easily generalized to measure the temporal orders across three frames. A triplet match can be defined as $i<j<k$ and $l_i \prec l_j \prec l_k$ and $\hat{l}_i \prec \hat{l}_j \prec \hat{l}_k$.
The triplet score shares a similar equation as pairwise score but changes the pairwise match to triplets.

Utilizing this approach, we are capable of procuring evaluation metrics of arbitrary length accuracy, such as when $\beta=\{2,3\}$, which yield \textit{pairwise score} and \textit{triplet score} respectively. In this paper, we primarily use the pairwise score because of its significant representativeness.

\paragraph{Human Evaluation} Lastly, due to the potential subjectiveness of movie reordering (see \cref{fig:qual_comp_2} for an example), the numerical comparison between the predicted order and ground truth may not always be the best metric. We apply human evaluation on predictions using ``Real \textit{vs.} Fake'' protocols; refer to ~\cref{sec:human-study} for details.

\subsection{HCMC Model}\label{sec:HCMC}

\begin{figure*}[t]
    \centering
    \includegraphics[width=.8\linewidth]{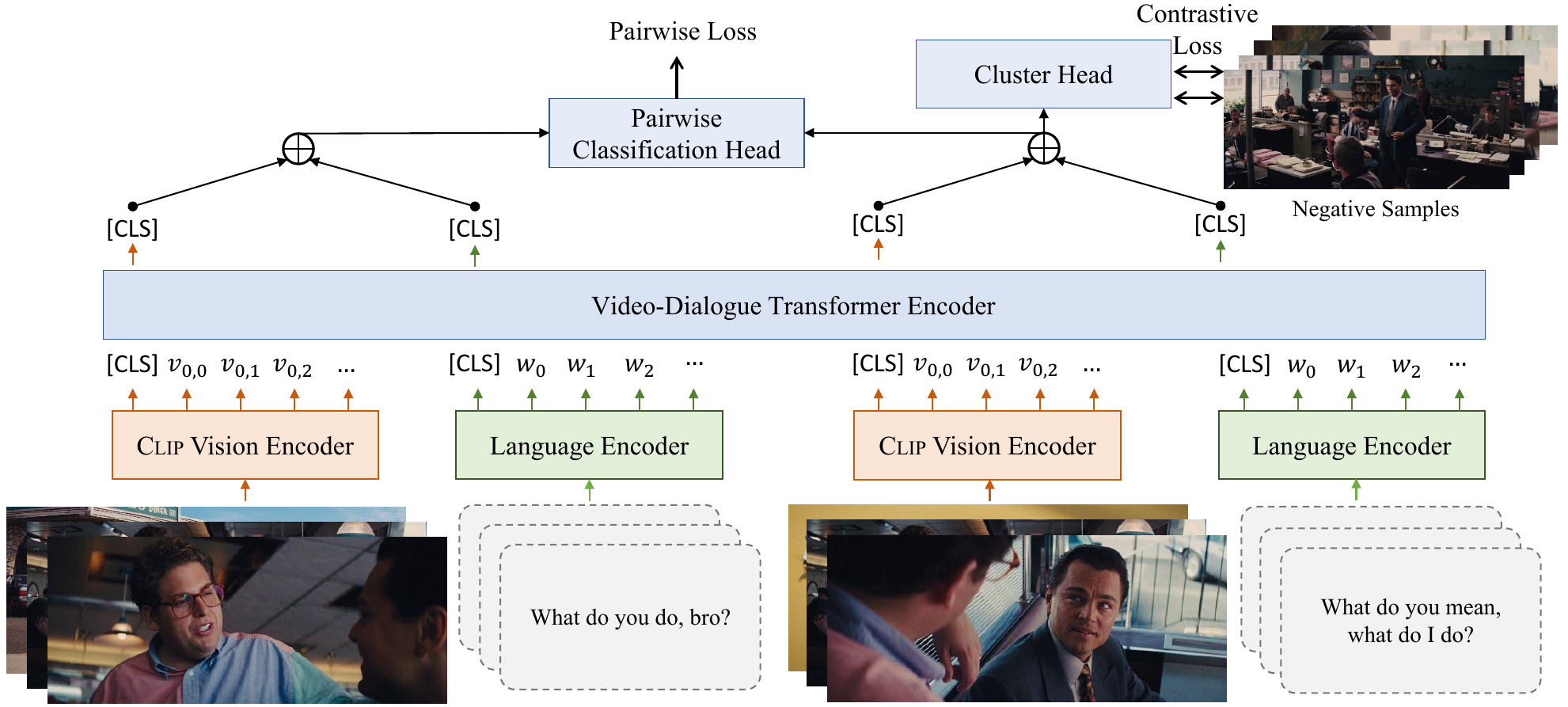}
    \caption{\textbf{An overview of frame-level \HCMC~model.}} 
    \label{fig:training}
\end{figure*}

The Hierarchical Contrastive Movie Clustering\sloppy(\HCMC) model is composed of separate language and image encoders, video-dialogue transformer encoders, pairwise classification heads, and cluster heads.
We jointly optimize our model with three-level ordering and two-level clustering tasks. An overview of \HCMC is sketched in \Cref{fig:training}. 

\paragraph{Multimodal Feature Extractor}\label{subsec:feature_extractor} 

Most existing video-language models\cite{zellers2021merlot, sharma2020deep, epstein2021learning} struggle to capture long-range temporal dependencies. To address this limitation, we adopt an end-to-end framework and use a CLIP-based encoder to extract visual and textual features from each movie clip. The video frames and their corresponding utterances are separately encoded into token sequences, which are then concatenated to form a unified frame-level multimodal representation that captures both appearance and dialogue context.

To reflect the hierarchical nature of cinematic content, we further aggregate these frame-level embeddings. Frames within a shot are combined to form shot-level features, and shots within a scene are further merged to construct scene-level representations. This structured representation enables the model to learn narrative patterns across multiple levels of abstraction.

\paragraph{Video-Dialogue Transformer Encoder}
After obtaining the embedding of frame and dialogue, we jointly optimize our model with a temporal ordering learning task and a contrastive learning task. For the ordering task, we first concatenate two randomly sampled image-dialogue pairs in the same level and then feed them to the frame-dialogue transformer encoder, after that we use a binary classification head to learn if the input is in the true temporal order. 

Notably, to mitigate exposure error, we sample frame-level pairs not only from the same shot. For the clustering task, we sample image-dialogue pairs from different groups as negatives, then optimize the cluster head by attracting the positive and repelling the negative samples.

\paragraph{Learning Objective}
We randomly sample an ordered pair of the same layer clips ($\mathcal{P}_i, \mathcal{P}_j, \widetilde{\mathcal{Q}}$ )  from the extracted embeddings, where $\mathcal{P}_i$ and $\mathcal{P}_j$ are two random embeddings belonging to the same level (depending on the layer in which the joint training occurs, the representation can be attributed to the frame, shot, or scene).~\review{\sout{R3: what are $P_i$, $P_j$, the input to the model is frame, shot or scene?}} and $\widetilde{\mathcal{Q}} = \{\widetilde{\mathcal{Q}}_1, \widetilde{\mathcal{Q}}_2, \cdots,\widetilde{\mathcal{Q}}_{n}\}$, where $n$ is the number of negative samples,  is the negative data sampled outside the current permutation group.
The pair-wise layer representation $(\mathcal{P}_i, \mathcal{P}_j)$ is fed into a binary classifier $\phi$, with two classes indicating the forward order ($\mathcal{P}_i \succ \mathcal{P}_j$) and the backward order ($\mathcal{P}_i \prec \mathcal{P}_j$) of input pairs. The cross-entropy loss is calculated based on the classifier output. Formally, the objective can be denoted as:
\begin{equation}
% \small
    \begin{aligned}
    \mathcal{L}_{cls} (\mathcal{P}_i, \mathcal{P}_j) = CE(\phi(\mathcal{P}_i,\mathcal{P}_j), O),
    \end{aligned}
\end{equation}
where $O \! \in \! \{0,1\}$ indicates the ground-truth ordering class and $CE$ stands for cross-entropy loss.

To derive a shot and scene-aware representation, we leverage the Contrastive Learning trick by feeding the output representation to another multi-class classifier $\psi$, where positive and negative data are distinguished based on the contrastive objective:
\begin{equation}
\begin{aligned}
\mathcal{L}_{CL} (\mathcal{P}_i, \mathcal{P}_j, \widetilde{\mathcal{Q}}) = 
 \!-\!\log\frac{e^{\psi(\mathcal{P}_i) \cdot \psi(\mathcal{P}_j)}}{e^{\psi(\mathcal{P}_i) \cdot \psi(\mathcal{P}_j)}\!+\!\sum_{k=0}^{n-1}e^{\psi(\mathcal{P}_i) \cdot \psi(\widetilde{\mathcal{Q}}_k)}},
\end{aligned}
\end{equation}
where $\phi(\mathcal{P}) \in \mathbb{R}_+$ is the representation generated by binary classifier, $\psi(\mathcal{P}) \in \mathbb{R}_+$ is the representation generated by multi-class classifier. 

Taken together, our final objective function is:
\begin{equation}
    \mathcal{L} = \mathcal{L}_{cls} + \lambda \mathcal{L}_{CL},
    \label{eqn:loss}
\end{equation}
where $\lambda$ is a balancing factor for two objectives.

\subsection{Top-down and Bottom-up  Inference}
We adopt a top-down clustering and bottom-up reordering pipeline approach for the inference. 

\paragraph{Reordering Algorithm}
For the training data set $X = \{x_{1}, \cdots , x_{n}\}$, there exists a confidence score between each pair of instances, represented by the weighted adjacency matrix ${\mathcal{S}} \in \mathbb{R}^{n\times n}$ where ${\mathcal{S}}[i,j]$ denotes the weight that $x_i$ is placed directly before $x_j$ in the predicted sequence. We use~\Cref{eqn:score_metric} to compute the score adjacency matrix: 
\begin{equation}
{\mathcal{S}}[i,j]\!=\! \frac{\exp{\phi(x_i, x_j)[1]-\exp{\phi(x_i, x_j)[0]}}}{\exp{\phi(x_i, x_j)[1]+\exp{\phi(x_i, x_j)[0]}} }.
\label{eqn:score_metric}
\end{equation}
Our goal is to find a path passing through all $n$ points in the complete graph such that the weight of the path is maximized. We utilize beam search to track the top $bsize$ paths at each iteration to search for local optimal solutions. The algorithm is described in supplementary material.

\paragraph{Top-down and Bottom-up Inference}
For each input test data, after going through the multimodel feature extractor, the input feature representation $X=\{ x_1, x_2, .. x_{N_{fm}}\}$ is obtained.
We run a top-down and bottom-up inference to obtain the final ordered sequence. 
We firstly adopt the top-down strategy for coarse-to-fine clustering scenes and then shots, and then we reorder the different levels of videos from bottom to up. Finally, we obtain the predicted temporal ordering sequence; see Supplementary Material for the full algorithm.

\section{Experiments}\label{app:exp}

\subsection{Movie Reordering}

\paragraph{Setup}
% We run all benchmark experiments using a single Nvidia 3090Ti.
The model is optimized using AdamW~\cite{oord2018representation} with learning rate as $1e-4$ and the following hyperparameters: $\beta_1 = 0.9$, $\beta_2 = 0.999$, $\epsilon = 1e-6$. All models are close to convergence with 5 epochs of training; refer to \textit{SM} for more implementation details.

\paragraph{Baselines} We take different types of prevalent pre-trained temporal models as baselines: language model ~(\Bert~\cite{kenton2019bert}, \AlBert~\cite{albert}, \ChatGPT~\cite{openai2023gpt4}), video model (\VideoMAE~\cite{tong2022videomae}, \Timesformer~\cite{DBLP:conf/icml/BertasiusWT21}), and video-language model (\Singularity~\cite{lei2022revealing}). In practice, we follow \Bert's next sentence prediction~(NSP) task and \AlBert's sentence-order prediction~(SOP) task to predict the temporal relation of the input pair by comparing the probability of binary classification. In addition to encoder-only models, we also incorporate recent advanced large video language models, including \VideoLLaVA~\cite{Lin2023VideoLLaVALU}, \LLavaNext~\cite{liu2024llavanext}, \TimeChat~\cite{ren2024timechat}, and \Vtimellm~\cite{huang2024vtimellm}.

\paragraph{Main Results}  
\begin{table*}[t!]
    \centering
    \small
    \captionof{table}{\textbf{Results on movie reordering.} Row 1 is a random baseline. Rows 2-4 are language-only models. Rows 5-6 are visual-only models. The rest are multi-modal performance. ``fusion'' indicates the early-fusion mechanism. *Traditional metrics could not be applied to the result from the generative model, because the results miss sequence numbers. not missing: indomain 49.75\% outdomain 48.58\% .}
    \label{tab:main_res}

    \resizebox{\linewidth}{!}{%
    \begin{tabular}{l|cccccc|cccccc}
    \toprule
        & \multicolumn{6}{c|}{in-domain} & \multicolumn{6}{c}{out-domain} \\
         & LSD $\downarrow$ & LMD $\downarrow$ & SD $\downarrow$ & SSD $\downarrow$ & OS ($\beta=2$) & OS ($\beta=3$)  & LSD $\downarrow$ & LMD $\downarrow$ & SD $\downarrow$ & SSD $\downarrow$ & OS ($\beta=2$) & OS ($\beta=3$) \\ \midrule
       Random  & 34.24 & 4.57 & 10.59 & 33.99 &49.66 &16.42  & 33.43 & 4.53 & 10.39 & 33.17 &49.75 &16.59 \\ \midrule
       \Bert & 32.91 & 4.41 & 10.38 & 32.83 &51.47 & 18.57 & 29.57 & 4.19 & 9.87 & 29.82& 52.16 & 18.79 \\
       \AlBert & \textbf{28.14} & \textbf{4.08} & 10.38 & 31.80 & 54.21 & 20.18 & 28.14 & 4.08 & 9.82 & 29.13 & 53.70 & 19.76 \\
       \GPT-3.5-turbo* & - & - & - & - & 51.80 & 21.25 & - & - & - & - & 53.34 & 22.78\\ 
       \GPT-4* & - & - & - & - & 53.18 & \textbf{24.53} & - & - & - & - & 53.69 & \textbf{24.87} \\ \midrule 
       \VideoMAE  & 33.66 & 4.54 & 10.54 & 33.80 & 50.31 & 17.71 & 30.23 & 4.29 & 9.56 & 30.39 & 50.78 & 18.19 \\ 
       \Timesformer  & 33.15 & 4.49 & 10.47 & 33.37 & 50.41 & 17.25 & 32.73 & 4.50 & 10.17 & 32.37 & 48.98 & 16.51 \\ \midrule
       \VideoLLaVA* & - & - & - & - & 35.66	& 11.72	& - & - & - & - & 38.36	& 12.64 \\
       \LLavaNext* & - & - & - & - & 30.05	& 9.76	& - & - & - & - & 31.7	& 9.92 \\
       \TimeChat* & - & - & - & - & 11.97	& 4.01	& - & - & - & - & 9.95	& 3.34 \\
       \Vtimellm*	& - & - & - & - & 29.63	& 9.85	& - & - & - & - & 32.87	& 10.95 \\
       \midrule
       \Bert+fusion & 32.02 & 4.33 & 10.29 & 32.50 &53.01 & 19.44 & 28.92 & 4.15 & 9.83 & 29.65 & 53.18 & 19.25 \\
       \AlBert+fusion & 30.75 & 4.25 & 10.41 & 31.83 & 54.06 & 20.30 & 28.23 & 4.09 & 9.92 & 29.18 & 53.70 & 19.76 \\ 
       \VideoMAE+fusion  & 33.32 & 4.50 & 10.48 & 33.38 & 50.67 & 18.55 & 30.40 & 4.30 & 9.93 & 30.50 & 50.83 & 18.67 \\
       \Singularity+fusion & 32.48 & 4.39 & 10.38 & 32.69 &51.75 & 18.67 & 29.75 & 4.22 & 9.88 & 30.00 & 52.33 & 18.86 \\ 
       \HCMC(frame+shot) & 29.64 & 4.11 & \textbf{10.18} & \textbf{30.72} & \textbf{55.40} & 21.58 & \textbf{27.29} & \textbf{3.98} & \textbf{9.82} & \textbf{28.41} & \textbf{54.97} & 21.23 \\
       \HCMC(frame+scene) & 30.36 & 4.23 & 10.37 & 31.66 &54.93 & 21.43 & 30.28 & 4.22 & 10.37 & 31.59 & 54.66 & 20.67 \\
       \HCMC(frame+shot+scene) & 31.69 & 4.34 & 10.40 & 32.37 &53.10 & 19.34 & 28.99 & 4.16 & 9.84 & 29.60 & 52.84 & 19.13 \\ \bottomrule
    \end{tabular}%
    }
\end{table*}

Tab.~\ref{tab:main_res} shows the main results of movie ordering on MoviePuzzle. We summarize our key observations as follows:
% and \cref{fig:frame_length} as follows:
\textbf{First, }\textit{temporal understanding predominantly rooted in dialogue continuity.} The robust performance of pure language models like \Bert
and \AlBert suggests that film sequence understanding largely depends on dialogue continuity, underscoring the importance of linguistic context. 
\textbf{Second, }\textit{vision models lack temporal support in image processing methods.} Vision models' performance, falling short of pure language models, points to a critical deficiency in current image processing methodologies: the lack of adequate support for temporality. \textbf{Third, }\textit{multimodal models require a greater focus on visual narrative structures.} Multimodal models like \Singularity, while showing promise, are yet to fully tackle the task. These models, preoccupied with image and text prediction, underscore the need for a renewed focus on visual narrative structures.
\textbf{Fourthly, }\textit{granularity of \textit{shot} structure boosts model performance.}  When compared, the \HCMC (frame+shot) model performs better than the \HCMC (frame+scene) model. This improvement points to the superior granularity provided by the \textit{shot} structure, particularly when a clip does not contain multiple scenes. Furthermore, by applying different video lengths, on MoviePuzzle, we find \HCMC surpassing all other models when the number of frames is more than $5$, indicating the necessity in visual narrative learning, detailed results can be found in Appendix.
\textbf{Lastly, } \textit{large video language models show poor zero-shot visual narrative reasoning ability}. We believe this comes from the lack of sophisticated data in the training phase of these models. In the following section, we will devise experiments to reveal the influence of temporal ordering training.

\begin{table}[t!]
\begin{minipage}[b]{0.48\textwidth}
    \centering
    \caption{\textbf{Ablated \HCMC~models.~\review{\sout{R3: Ablation study should also show the results without frames or scenes}}}}
    \label{tab:Ablation-Study}
    \resizebox{\linewidth}{!}{%
    \begin{tabular}{lcccc}
    \toprule
        & \multicolumn{2}{c}{in-domain} & \multicolumn{2}{c}{out-domain} \\ \cmidrule{2-5}
         & $\beta=2$ & $\beta=3$ & $\beta=2$ & $\beta=3$ \\ \midrule
       \HCMC(frame+shot) &\textbf{55.40} & \textbf{21.58} & \textbf{54.97} & \textbf{21.23} \\
       w/o shot & 54.21 & 20.41 &  54.29 & 20.19 \\
       w/o CL & 52.63 & 18.79 & 52.74 & 18.05 \\
       w/o text & 50.05 & 17.53 & 49.86 & 17.40 \\
       w/o vision & 51.69 & 18.36 & 51.87 & 18.67 \\ \bottomrule
    \end{tabular}
    }
\end{minipage}
\hfill
\begin{minipage}[b]{.5\textwidth}
    \centering
    \caption{\textbf{Accumulation error analysis.}}
    \label{tab:acc_error}
    \resizebox{\linewidth}{!}{%
    \begin{tabular}{lccccccc}
    \toprule
        \multicolumn{2}{c}{}& \multicolumn{3}{c}{in-domain} & \multicolumn{3}{c}{out-domain} \\ \cmidrule{3-8}
        & & IoU & $\beta=2$ & $\beta=3$ & IoU & $\beta=2$ & $\beta=3$ \\ \midrule
         \multirow{2}{*}{Euclidean} &shot & \textbf{37.56} & 54.93 & 21.41 & \textbf{35.74} & 54.98 & \textbf{21.33} \\ &frame & - & \textbf{55.40} & \textbf{21.58} & - & 54.97 & 21.23\\ \midrule
         \multirow{2}{*}{Cosine} &shot & 36.9 & 54.42 & 20.49 & 36.70 & \textbf{55.17} & 21.31\\&frame & - & 54.77 & 21.15 & - &  55.10 & 21.28\\ \midrule
         \multirow{2}{*}{Soft\_DTW} &shot & 16.94 & 52.71 & 18.04 & 16.85 & 53.49 & 19.83\\&frame & - & 53.24 & 19.94 & - & 54.70 & 20.94\\ \bottomrule
    \end{tabular}
    }
\end{minipage}
\end{table}

\subsection{Ablation Studies}\label{sec:analysis}

\paragraph{HCMC Components} Tab.~\ref{tab:Ablation-Study} shows the ablation experiments of different components in the \HCMC model. The accuracy of the experiments decreases when any layer shown in the table is removed. Among them, removing the textual information in the clip has the most significant impact on the model, followed by the image information. It can be seen that the \HCMC relies heavily on multimodal information, which is consistent with our intuition. When contrastive learning or the shot layer is removed from the model, the accuracy of the model also decreases.

\paragraph{Accumulative Error}
Our three-tier fine-grained model with both shot and scene clustering modeling obtains poor performance. We believe the accumulative error from the clustering layer and re-order layer causes this. Thus we quantitatively analyze the intermedia result of different stages. Tab.~\ref{tab:acc_error} shows the inter-media score of the shot clustering and frame ordering among different distance functions. We calculate the Intersection-over-Union~(IoU) score of the shot clustering for reference. Specifically, we match the cluster centers with the mean of the ground truth cluster through cosine similarity. The results demonstrate the shot clustering score is positively correlated with the shot ordering score. However, the  drop ordering score is more mitigated than the clustering score. This is because the soft\_DTW~\cite{Cuturi2017SoftDTWAD} method results in more empty clusters.

% \subsection{How does ordering affect holistic video understanding?}\label{sec:msa}
\subsection{Downstream video understanding}\label{sec:msa}

\begin{table}[t!]
\begin{minipage}[b]{0.55\textwidth}
    \centering
    % \small
    \caption{\textbf{Results on Movie Synopsis Association.} ``Reordering" refers to the model with our video reordering task added upstream.}
    \label{tab:synopsis}
    \resizebox{\linewidth}{!}{%
    \begin{tabular}{lcccccccc}
    \toprule
        & R@1($\uparrow$) & R@5($\uparrow$) & R@10($\uparrow$) & MedR($\downarrow$) \\ \midrule
       Random & 0.13 & 0.66 & 1.32 & 378.5  \\ \midrule
       zero-shot & 0.26 & 1.06 & 2.25 & 230.5 \\
       zero-shot+reordering & \textbf{0.40} & \textbf{1.46} & \textbf{3.04} & \textbf{220.0} \\ \midrule
       \Singularity~\cite{lei2022revealing} & \textbf{10.19} & 30.16 & 39.95 & 18.5   \\
       \Singularity+reordering & 9.92 & \textbf{31.88} & \textbf{43.39} & \textbf{14.0} \\ \midrule
       % \VideoLLaVA & 18.0 & 38.0 & 52.0 & 9.5 & 22.0 & 52.0 & 66.0 & 4.0  \\
       \VideoLLaVA~\cite{Lin2023VideoLLaVALU} & 3.57 & 9.79 & 15.08 & \textbf{104.0}   \\
       \VideoLLaVA+reordering &\textbf{3.84} & \textbf{10.32} & \textbf{16.14} & 109.0     \\ \bottomrule
    \end{tabular}%
    }
\end{minipage}
\hfill
\begin{minipage}[b]{0.42\textwidth}
    \centering
    % \small
    \caption{\textbf{Results on MLVU.} We selected sub-tasks related to temporal understanding from the task set of MLVU.} 
    %In this context, TR stands for Topic Reasoning, AR refers to Anomaly Recognition, VS represents Video Summarization, and AO denotes Action Order.}
    \label{tab:MLVU}
    \resizebox{\linewidth}{!}{%
    \begin{tabular}{lcccc}
        \toprule
        % &Topic Reasoning	&Anomaly Recognition	&Video Summarization	&Action Order \\ \midrule
        &TR	&AR	&VS	&AO \\ \midrule
        Full Mark & 100 &100 &10 &100 \\ \midrule
        \VideoLLaVA	&71.6	&57.0	&2.43	&20.1 \\
        \VideoLLaVA + reordering	&\textbf{74.3}	&\textbf{57.1}	&\textbf{2.83}	&\textbf{29.8} \\ \midrule
        \TimeChat	&23.1	&\textbf{27.0}	&2.54	&24.7 \\
        \TimeChat + reordering	&\textbf{29.4}	&25.9	&\textbf{2.81}	&\textbf{28.0} \\
        \bottomrule
    \end{tabular}%
    }
\end{minipage}
\end{table}

\begin{figure*}[htbp]
    \centering
    \includegraphics[width=.95\linewidth]{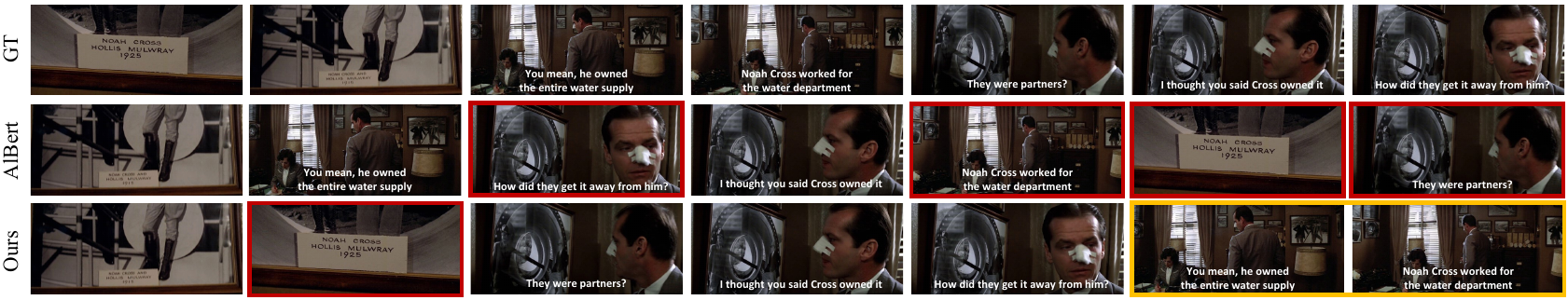}
    \caption{\textbf{Qualitative comparisons on reordering performance between \AlBert+Fusion~and \HCMC~(Ours).} The red rectangles highlight misplaced frames, while the yellow ones denote misplaced shots (with correct inner frame orders).
    Our models excel in grouping together frames that share similar shot information, thereby forming a more plausible visual narrative.}
    \label{fig:qual_comp}
\end{figure*}

\begin{figure*}[htbp]
    \centering
    \includegraphics[width=.95\textwidth]{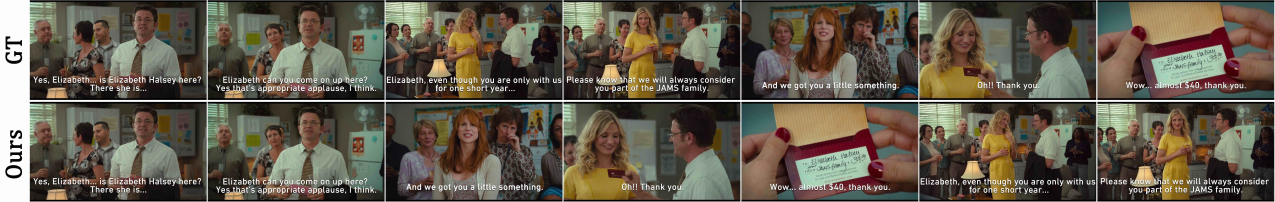}
    \caption{\textbf{Qualitative comparisons between Ground Truth and \HCMC(Ours).} In this context, the temporal sequence of Elizabeth stepping forward to receive the gift and the male host delivering a speech can be considered interchangeable. The \HCMC model demonstrates a rational interpretation of the narrative.}
    \label{fig:qual_comp_2}
\end{figure*}

Although reordering has been validated as an effective proxy task for short videos~\cite{epstein2021learning, yeung2016end, DBLP:journals/corr/MisraZH16, DBLP:journals/corr/abs-2004-02205}, we evaluate its impact on holistic understanding of long-form videos through the Movie Synopsis Association task~\cite{xiong2019graph}. This task requires the model to retrieve the correct text synopsis for a given movie segment, demanding strong alignment across multimodal semantic units.

We measure performance using standard retrieval metrics Recall@N (R@N) and Median Rank (MedR). As shown in Table~\ref{tab:synopsis}, despite the simplicity of the reordering task~\cite{zellers2021merlot}, we observe consistent gains from reordering and scene-level embeddings. This highlights temporal ordering as a crucial component for bridging visual and textual semantics in long-form narrative understanding.

We further assess downstream generalization on the MLVU benchmark~\cite{zhou2024mlvucomprehensivebenchmarkmultitask}, including Topic Reasoning (TR), Anomaly Recognition (AR), Video Summarization (VS), and Action Order (AO). TR, AR, and AO are multiple-choice tasks, while VS is evaluated using GPT-4 by comparing generated summaries with annotations. Results in Table~\ref{tab:MLVU} confirm that pretraining with the reordering task leads to significant performance improvements across these diverse video understanding challenges.

\begin{wraptable}{r}{0.55\textwidth}
    \centering
    \caption{\textbf{Human preferences.}
Consider incorporating human preferences in the understanding task.}
    \resizebox{\linewidth}{!}{%
    \begin{tabular}{l|ccc}
        \toprule
       & all & in-domain & out-domain \\
       \midrule
       \HCMC (Ours) v.s. \textsc{AlBert}  &0.55 & 0.54 & 0.56  \\
       \HCMC (Ours) v.s. Ground Truth & 0.26 & 0.23 & 0.30  \\
       \bottomrule
    \end{tabular}
    }
    \label{tab:human}
\end{wraptable}

\subsection{Human Study}\label{sec:human-study}

To verify our model's inherent understanding of movie logic, we recruited 10 well-educated human testers with proficient English comprehension skills to compare the logical coherence of generated clip sequences. Specifically, we select 20 sets of test data from both in-domain and out-domain sources and fed them into the best-performing baseline model \AlBert, our \HCMC~model and ground truth for prediction. \Cref{fig:qual_comp} showcases a random example. Even when the \AlBert model misplaces frames, our model maintains accuracy within the same shot, and testers, presented with randomly shuffled predicted sequences, consistently choose the more reasonable one without knowing the source.
The comparison results in~Tab.~\ref{tab:human} demonstrate that the sequences generated by our approach are more in line with human preferences with \AlBert. 
In comparison to the ground truth, there is approximately a 26.5\% likelihood for testers to select the images generated by our model. This is because there are often multiple plausible scenarios for rational frame reordering in films, as exemplified in the provided image~\Cref{fig:qual_comp_2}.More results can be find in Appendix.
% ~\Cref{sec:qualitative_image}.

\section{Related Work}
\iffalse
\paragraph{Joint representation of images and text}
Joint image-text representations~\cite{bengio2013representation} benefit many language-and-vision tasks by fusion of the modalities.
A family of ``VisualBERT'' models~\cite{tan2019lxmert, chenuniter, alberti2019fusion, li2019visualbert, lu2019vilbert, yu2021ernie, gan2020large} have been proposed a common method which uses a supervised object detector image encoder backbone and pre-train on image-caption pairs. Cross-modal representations are learned through masked language modeling objective~\cite{kenton2019bert}
Another family of vision-language models is based on contrastive learning~\cite{alayrac2020self,bain2021frozen,li2021align, alec2021clip,yao2021filip}, which have a solid ability to extract features on static image-caption pairs.
Some other models like Flamingo~\cite{alayrac2022flamingo} and Blip-v2~\cite{li2023blip} use a lightweight transformer to bridge the modality gap between a frozen image encoder and a frozen large language model~(LLM). 
Our method differs as it uses an explainable image-text representation that remains unchanged when learning different semantic structures.
\fi

\paragraph{Video-Language Understanding}
As an application of artificial intelligence in the multi-media field, video-language understanding has drawn great attention in the research community, such as video story telling~\cite{huang2016visual}, video moment retrieval~\cite{chen2020fine}, image caption~\cite{hou2020joint, yang2019auto}, visual question answering~\cite{li2019relation}, and action recognition~\cite{wang2018non}. Prior arts before the large-scale pre-training era~\cite{lei2018tvqa, lei2019tvqa, le2020hierarchical} leverage offline extracted video features~\cite{kay2017kinetics}, after that, video-language pre-trained models~\cite{li2020hero, zhu2020actbert} have shown promising results. Aligned with the success of transformer-based~\cite{vaswani2017attention} language pre-training models~\cite{liu2019roberta, yang2019xlnet}, image-text pre-training~\cite{li2019visualbert, he2022masked} and video-text pre-training~\cite{kim2021self, fu2023empirical-mvm} usually use masked visual modeling and have shown promising results on short videos clips. 
These models underutilized temporal information in long videos~\cite{zellers2021merlot, huang2020movienet}. Larger models, such as VideoLLaVA~\cite{Lin2023VideoLLaVALU}, VideoChat~\cite{2023videochat}, and TimeChat~\cite{ren2024timechat}, have also achieved strong performance.

\paragraph{Narrative Video Benchmarks}
A couple of research works have addressed the importance of understanding long-form videos, especially movies~\cite{lsmdc, ava, laptev2008movie-act,Bha2014cls}.
For example, MovieQA~\cite{movieqa} is extracted from 408 movies with 15K questions. This dataset is designed by using QA to evaluate story understanding.
MovieGraph~\cite{moviegraphs} is a small dataset with graph-based annotation of social relationships depicted in clips edited from 51 movies. With the relation graph of character, interaction, and attributions, MovieGraph can offer a hierarchical structure of movie understanding. Some other video datasets, such as LongVideoBench~\cite{wu2024longvideobench}, MLVU~\cite{MLVU}, Video-MME~\cite{video-mme}, and Shot2Story~\cite{han2023shot2story20k}, also include longer movie data but not as their primary focus.
MovieNet~\cite{huang2020movienet} is a comprehensive movie dataset with diverse annotations for in-depth understanding. From its 1,100 movies, we filter and organize well-labeled, aligned subsets within a movie hierarchy.

\section{Conclusion}\label{sec:conclusion}

We present MoviePuzzle, a new benchmark designed to evaluate visual narrative reasoning by challenging models to reconstruct coherent storylines from shuffled multimodal clips. To support this, we curate a hierarchically annotated dataset and propose the HCMC model, which leverages structural cues for temporal reordering. Our Ordering Score further enables precise evaluation of narrative coherence. Experiments show that our approach improves long-form video understanding and opens new directions for automated video analysis and editing.

\clearpage
\bibliographystyle{unsrt}
\bibliography{reference}

%%%%%%%%%%%%%%%%%%%%%%%%%%%%%%%%%%%%%%%%%%%%%%%%%%%%%%%%%%%%
\clearpage

% 

%%%%%%%%%%%%%%%%%%%%%%%%%%%%%%%%%%%%%%%%%%%%%%%%%%%%%%%%%%%%

% \input{check_list}
\appendix

\begin{center}
    \textit{\Large Supplementary Materials}
\end{center}

% \appendix

% \section{Appendix}

We provide supplementary materials as follows:

\begin{itemize}[noitemsep, leftmargin=25pt, topsep=2pt]
    % \item \Cref{sec:related_work} is the related work.
    \item In~\Cref{sec:data_collection}, we describe the steps we took to collect and preprocess our dataset.
    \item In ~\Cref{sec:OS-metric}, we present a detailed description of our new proposed Ordering Score.
    \item \Cref{sec:inference_algorithm} provide a detailed description of our inference algorithm.
    \item \Cref{sec:impl_detail} presents the experimental hyperparameters and implementation details of our main experiment.
    \item \Cref{sec:alternative_design} introduce alternative model designs and presented experimental findings.
    \item In section~\Cref{sec:more_numerical_results}, we investigate the impact of clip length on experimental results.
    % \item \Cref{sec:human_study} conducted an additional human study experiment.
    % \item In~\Cref{sec:ethics_concern} we discuss potential ethical risks associated with our paper.
    \item \Cref{sec:limitation} shows our limiation
    \item \Cref{sec:qualitative_image} displays some qualitative result images.
    % \item \Cref{sec:datasheets} conducts datasheets~\cite{gebru2021datasheets} of our \MoviePuzzle dataset.
    \item \Cref{sec:license} presents our responsibility and license.

\end{itemize}

\section{Data Collection Information}\label{sec:data_collection}

\subsection{Selecting Movie}
We employ the following top-level criteria for choosing the movie source and its associated transcripts from MovieNet
%here ~\cite{huang2020movienet}
for \MoviePuzzle:
\begin{enumerate}[leftmargin=20pt]
\item The movies should have color images, and the quality of the picture should not be too dark or blurry.
\item The movies should have English subtitles, as some movies in the original MovieNet dataset either have missing subtitles or have subtitles in languages other than English.
\item The movies should have synopsis summaries to assist in downstream Movie Synopsis Association tasks. However, it should be noted that only a subset of movies in the original MovieNet dataset have synopsis labels.
\item The movies should have a list of actors. We plan to add character labels to the dialogue in future work.
\item The movies should have defined boundaries for shots and scenes.
\end{enumerate}

\subsection{Aligning Subtitle}
We match each frame image with its corresponding subtitle using the following steps:
\begin{enumerate}[leftmargin=20pt]
\item Firstly, we remove most voiceovers and background sounds from the subtitles and replace foreign words with English letters with tones.
\item Secondly, each dialogue is only paired with images that fall exactly within its corresponding time period, rather than the closest ones.
\item Finally, in the case of multiple frames falling on a single dialogue, we merge these frames and select the middle frame as the most representative matching image.
\end{enumerate}

\subsection{Cutting into Clips}
We segment the movie into clips based on the following criteria, where the textual subtitles are aligned frame by frame:
\begin{enumerate}[leftmargin=20pt]
\item The length of each clip is restricted to 10-20 frames.
\item The frames containing subtitles in each clip must account for more than 80\% of all.
\item A greedy algorithm is utilized to match the movie for as long as possible.
\end{enumerate}

After obtaining a total of 10,031 movie clips, we split the data into training, validation, in-domain test, and out-of-domain test sets in the proportions of 70\%, 6\%, 12\%, and 12\%, respectively. The dataset will be publicly released at a later date.

\section{New Proposed Metric: Ordering Score}~\label{sec:OS-metric}
Traditional metrics for sequence similarity include \textbf{LSD}, \textbf{LMD}, \textbf{SD}, and \textbf{SSD}. However, these metrics necessitate the number of predicted sequences to be the same as the reference, rendering them inapplicable to the results of a generative model which might miss some sequences. Therefore, we propose a new metric for temporal sequence evaluation.

% As defined in \Cref{sec:task-form}, 
The video clip has $N_{fm}$ frames, the ground truth sequence is $\{ l_1, l_2, \cdots, l_{N_{fm}}\}$, and model ordering prediction sequence is $\{\hat{l}_1, \hat{l}_2, \cdots, \hat{l}_{N_{fm}}\}$. We define a match as a $\beta$ length subset of elements that follow their natural order: $i_1 < i_2 < \cdots < i_\beta$ and $l_{i_1} \prec l_{i_2} \prec \cdots \prec l_{i_\beta}$ and $\hat{l}_{i_1} \prec \hat{l}_{i_2} \prec \cdots \prec \hat{l}_{i_\beta}$, where $\prec$ is a comparator indicating the precedent temporal order. The ordering score is computed as the ratio of the cardinality of the match set:

\iffalse
\begin{small}
\begin{equation}
    \text{OrderingScore} = \\
    \frac{\# \{(i_k)_{k=1}^{n} | (i_1 < \cdots < i_\beta) \wedge (l_{i_1} \prec \cdots \prec l_{i_\beta}) \wedge (\hat{l}_{i_1}  \prec \cdots \prec \hat{l}_{i_\beta}) \}} {{\tbinom {N_{fm}} \beta}},
    \label{eqn:ordering_score}
\end{equation}
\end{small}
where ${\tbinom {N_{fm}} \beta} = \frac{N_{fm}!}{\beta!(N_{fm}-\beta)!}$ is the total number of index combinations.
\fi

\begin{equation}
    \text{OrderingScore} = \frac{|\mathcal{M}|}{{\tbinom {N_{fm}} \beta}}
\end{equation}
where $\mathcal{M} = \{(i_k)_{k=1}^{n} | (i_1 < \cdots < i_\beta) \wedge (l_{i_1} \prec \cdots \prec l_{i_\beta}) \wedge (\hat{l}_{i_1}  \prec \cdots \prec \hat{l}_{i_\beta}) \}$ represents the index set of all sequences that satisfy the order matching., and ${\tbinom {N_{fm}} \beta} = \frac{N_{fm}!}{\beta!(N_{fm}-\beta)!}$ is the total number of index combinations.

\paragraph{Pairwise Score} The pairwise matching score is one of the most commonly used metrics for permutations in discrete mathematics and permutation graph theory.

\paragraph{Triplet Score} The pairwise score can be easily generalized to measure the temporal orders across three frames. A triplet match can be defined as $i<j<k$ and $l_i \prec l_j \prec l_k$ and $\hat{l}_i \prec \hat{l}_j \prec \hat{l}_k$.
The triplet score shares a similar equation as pairwise score but changes the pairwise match to triplets.

Utilizing this approach, we are capable of procuring evaluation metrics of arbitrary length accuracy, such as when $\beta=\{2,3\}$, which yield \textit{pairwise score} and \textit{triplet score} respectively. In this paper, we primarily use the pairwise score because of its significant representativeness.

\section{Inference Algorithm}\label{sec:inference_algorithm}
%-----
We adopt a top-down clustering and bottom-up pipeline approach for the reordering inference. Specifically, we adopt the Top-down strategy for coarse-to-fine clustering scenes and then shots, and then we reorder the different levels of videos from bottom to up. Suppose $X=\{ x_1, x_2, \cdots, x_{N_{fm}}\}$ is a list of shuffled video frames clip, with ground truth index set $\{ l_1, l_2, \cdots, l_n\}$. We complete the inference in five steps as follows and~\Cref{alg:inference}:

\paragraph{Frame cluster to scene}
We use the output of the scene-level cluster head of \HCMC~as the feature for k-means clustering. We can obtain the current scene layer clustering groups as 
$$
\hat{X}^{1}=\{g_k; g_k=\{x^k_1, \cdots, x^k_{N^k_{sn}}\}\}_{k=1}^{N_{sn}},
$$ 
where each group $h_k$ represents the index of the elements in the same scene.

\paragraph{Frame cluster to shot}
After clustering the scene groups as $H$, we use the output of the shot-level cluster head of \HCMC~as the feature for k-means clustering on shot. At this point, we obtain the grouping of the shot-level clustering as 
$$
\hat{X}^{2}=\{g_k; g_k = \{f^k_i, f^k_i = \{x^{k,i}_j \}_{j=1}^{N_{st}^{k,i}} \}_{i=1}^{N^k_{sn}} \}_{k=1}^{N_{sn}}.
$$

\paragraph{Frame-level reordering}
After grouping by shot, we can use the frame-level classification head of \HCMC~to sort the frames in each shot. We can obtain the output vector $p$ between any two frames using a binary classifier, and use the difference in softmax to represent the confidence of the order representation. In this way, we can obtain a weight value between any two frames, and we can obtain an adjacency list. We use beam search to search for the maximum weight arrangement, representing that we have sorted the order of all scenes on a clip using binary classification. At this point, the order becomes 
$$
\hat{X}^{3}=\{g_k; g_k = \{f^k_i, f^k_i = \{\hat{x}^{k,i}_j \}_{j=1}^{N_{st}^{k,i}} \}_{i=1}^{N^k_{sn}} \}_{k=1}^{N_{sn}}.
$$

\paragraph{Shot-level reordering}
After reordering frames for each shot, we sort the shots in each scene. Firstly, we use the concatenated features of each frame in the shot to represent the current shot feature. Then we use the shot-level classification head of \HCMC~to obtain an adjacency list representing the order confidence. We then use the same beam search method to find the ordering with the maximum total confidence. At this point, the order becomes
$$
\hat{X}^{4}=\{g_k; g_k = \{\hat{f}^k_i, \hat{f}^k_i = \{\hat{x}^{k,i}_j \}_{j=1}^{N_{st}^{k,i}} \}_{i=1}^{N^k_{sn}} \}_{k=1}^{N_{sn}}.
$$

\paragraph{Scene-level reordering}
Finally, we sort the scenes by pairing the features of these scenes, using the scene-level classification head of \HCMC~model to obtain an adjacency list representing the order confidence, and then using the same beam search method to find the ordering with the maximum total confidence. At this point, the scene layer has been ordered as 
$$
\hat{X}^{5}=\{\hat{g}_k; \hat{g}_k=\{\hat{x}^k_1, \cdots, \hat{x}^k_{N^k_{sn}}\}\}_{k=1}^{N_{sn}},
$$ 
Expanding it into a one-dimensional list $\hat{X}=\{\hat{x}_1, \cdots, \hat{x}_{N_{fm}}\}$ and gaining its index list represents the order of the prediction as $\{\hat{l}_1, \hat{l}_2, \cdots, \hat{l}_n\}$. 

% In conclusion, we employed the metric discussed in~\Cref{sec:metric} to compute the accuracy, specifically utilizing the $Pair/TripleScore$ formula.

%-----

% \begin{wrapfigure}{r}{.6\linewidth}
\begin{algorithm}
    \DontPrintSemicolon
    \SetNoFillComment

    \caption{Top-down and Bottom-up Inference}
    \label{alg:inference}
        \KwIn {temporally shuffled sequence $X$, classifier embedding layer $\phi$, clusterer embedding layer $\psi$, scene count $N_{sn}$, shot count ${N_{st}}$, beam search width $bsize$}
        \KwOut {temporally ordered sequence $Y$}
        \tcc {1. scene clustering }
        $X \gets KmeansClustering(\psi(X), N_{sn})$\;
        \tcc {2. shot clustering}
        $tmp \gets$ empty set \;
        \For{$scene$ in $X$}{
                $tmp$.add($KmeansClustering(\psi(scene),$
                ${N_{st}^{scene}}$)\;
        }
        $X \gets tmp$\;
        \tcc {3. frame reordering}
        \For{$scene$ in $X$}{
                \For{$shot$ in $scene$}{
                    ${\mathcal{S}} \gets$ substitute $shot$ into weighted adjacency matrix\;
                    $path \gets BeamSearch({\mathcal{S}}, bsize)$\;
                    Sort the $shot$ in $X$ based on the sequence $path$\;
                }
        }
        \tcc { 4. shot reordering }
        \For{$scene$ in $X$}{
            ${\mathcal{S}} \gets$ substitute $scene$ into weighted adjacency matrix\;
            $path \gets BeamSearch({\mathcal{S}}, bsize)$\;
            Sort the $scene$ in $X$ based on the sequence $path$.\;
        }
        \tcc {5. scene reordering }
        ${\mathcal{S}} \gets$ substitute $X$ into weighted adjacency matrix\;
        $path \gets BeamSearch({\mathcal{S}}, bsize)$\;
        Sort the $X$ based on the sequence $path$.\;
        % \STATE {Expanding $X$ into a one-dimensional sequence}
        \Return {$X$}\;
\end{algorithm}
% \end{wrapfigure}

\subsection{Kmeans Clustering Algorithm}
Suppose we have a training set $X = \{x_{1}, \cdots , x_{n}\}$, and want to group the data into a few cohesive clusters. Here, we are given feature vectors for each data point $x_{i}$ with no labels as an unsupervised learning problem. Our goal is to predict $m$ centroids and a label $z_{i}$ for each data point. 
Then cluster sequence $x$ into $m$ groups based on its corresponding $z_{i}$ to obtain a new sequence $Y = \{y_{1}, \cdots, y_{m}\}$, where $Y$ is a partition of $X$.
The kmeans clustering algorithm is as~\Cref{alg:kmeans}:

\begin{algorithm}
    \caption{KmeansClustering}
    \label{alg:kmeans}
    \DontPrintSemicolon
    % \begin{algorithmic}[1]
        \KwIn {training set ${X}$, number of clusters $m$, max number of steps $K$}
        \KwOut {The clustered sequence $Y$}
        Initialize cluster centroids $\mu_1, \cdots, \mu_m \!\in\!\mathbb{R}^m$ randomly\;
        \ForEach {$i$}{ 
            $z_{i} \gets {\rm arg}\mathop{{\rm min}}\limits _j \lVert x_{i} - \mu_{j} \lVert^2 $\;}
        \ForEach {$j$} {
            $\displaystyle \mu_j \gets \frac{\sum_{i=1}^{n} 1 \{z_{i} = j\}x_{i}}{\sum_{i=1}^{n} 1 \{z_{i} = j\} } $
        }
        Repeat step \textbf{2-7} until convergence or reaching max cluster steps\;
        $Y\!\gets X's$ partition that for all $x_{i} \in y_{j}$ have same $z_{i}$\;
        \Return {$Y$}\;
    % \end{algorithmic}
\end{algorithm}

\subsection{Beam Search Algorithm}
For the training data set $X = \{x_{1}, \cdots , x_{n}\}$, there exists a confidence score between each pair of instances, represented by the adjacency matrix ${\mathcal{S}}$ where ${\mathcal{S}}$ size is $n\times n$ and ${\mathcal{S}}[i,j]$ denotes the weight that $x_i$ is placed directly before $x_j$ in predict sequence. Our goal is to find a path passing through all $n$ points in the complete graph such that the weight of the path is maximized. We utilize beam search to keep track of the top $bsize$ paths at each iteration to search for local optimal solutions. The algorithm is described in~\Cref{alg:beam_search}.

\begin{algorithm}
    \DontPrintSemicolon
    \SetNoFillComment
    \caption{BeamSearch}
    \label{alg:beam_search}
    % \begin{algorithmic}[1]
        \KwIn {${\mathcal{S}}$(weighted adjacency matrix), $bsize$}
        \KwOut {$path$}
        {${bestlist} \gets empty$}\;
        {$n \gets$ length of ${\mathcal{S}}$}\;
        \For{$begin$ in $n$}{
            {${beamlist} \gets \{ begin\} $}\;
            {$len \gets 1$}\;
            \While{$len < n$}{
                {${newlist} \gets empty$}\;
                \For{$beamPath$ in ${beamlist}$}{
                {${newlist}$.add(\{$beamPath+i$, \textbf{for all} node $i$ not in $beamPath$\})}\;
                {${newlist} \gets$ top $bsize$ score path in ${newlist}$}\;
                }
                {${beamlist} \gets {newlist}$}\;
                {$len \gets len+1$}\;
            }   
            {${bestlist}$.add(max score path in ${beamlist}$)}\;
        }
        {$path \gets$ max score path in ${bestlist}$}\;
        \Return{$path$}\;
    % \end{algorithmic}
\end{algorithm}

\section{Additional Implementation Details}\label{sec:impl_detail}
\begin{figure}
    \centering
    \resizebox{\linewidth}{!}{\includegraphics{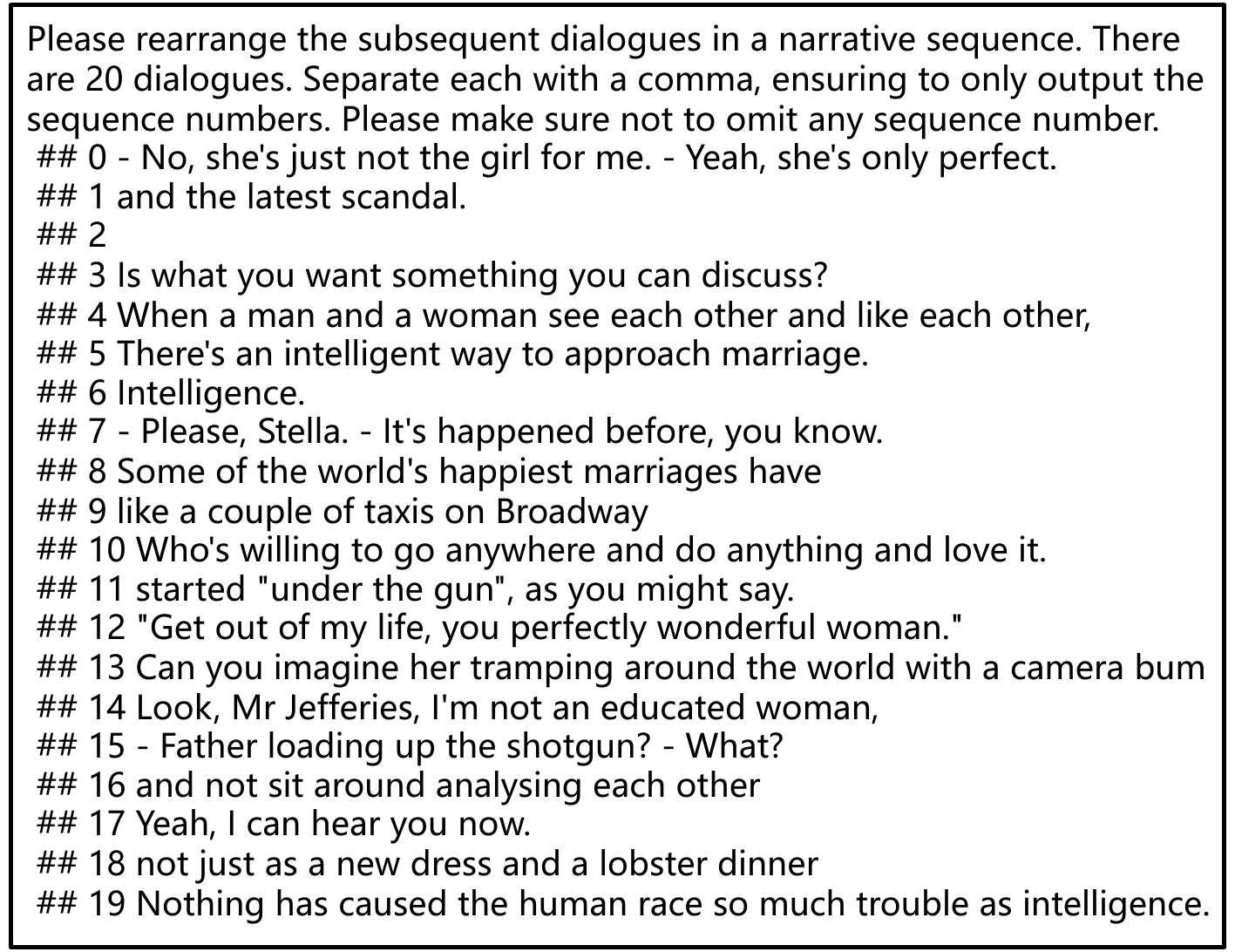}}
    \caption{Prompt for GPT-4}
    \label{fig:prompt}
\end{figure}

The hyper-parameters we use are shown in~\Cref{tab:Hyperparam} below. Our \Clip
%here ~\cite{alec2021clip}
(clip-vit-base-32\footnote{ \url{https://huggingface.co/openai/clip-vit-base-patch32}}) and \textsc{Bert}~\cite{kenton2019bert}~(bert-base-uncased\footnote{\url{https://huggingface.co/bert-base-uncased}}) parts code uses from HuggingFace. For more implementation details, please refer to \url{https://anonymous.4open.science/r/MoviePuzzle}. In addition, In ~\Cref{fig:prompt}, we present the prompt of our implementation of GPT4 baseline.

\begin{table}[ht]
\begin{center}
\caption{Hyperparameters for \HCMC.}
\begin{tabular}{lccc}
\toprule
\multicolumn{1}{l}{\bf Hyperparam} &\multicolumn{1}{c}{\bf HCMC} \\ 
\midrule 
Number of Transformer Layers  & 2  \\
Hidden Size & 512   \\
Attention Heads & 8 \\
Attention Heads Size & 64 \\
Learning Rate & 1e-4 \\
Batch Size & 8  \\
Max Cluster Steps & 1000 \\
Cluster Distance  & Euclidean \\
Epoch & 5 \\
AdamW $\epsilon$ & 1e-6 \\
AdamW $\beta_1$ & 0.9 \\
AdamW $\beta_2$ & 0.999 \\
Weight Decay & 0.01 \\
Patch Size & 32 \\
\bottomrule
\end{tabular}

\label{tab:Hyperparam}
\end{center}
\end{table}

\section{Alternative Design}\label{sec:alternative_design}
We have experimented with model structures trained separately at the frame, shot, and scene levels, in addition to the joint end-to-end training structure introduced in the main body of the paper. We also want to try layered experiments in addition to see how the combined training model can help.

In essence, we trained three parallel independent models, each having a similar structure. For the Video-Dialogue Encoder layer, we have tried using both a simple MLP structure and a simple transformer block. The result shows that these single-layer training models achieved satisfactory results in their respective layers (as shown in rows 1-3 of ~\Cref{tab:main_mlp,tab:main_transformer}). However, when employing the trained single-layer models for multi-layer inference (as shown in row 4-6 of ~\Cref{tab:main_mlp,tab:main_transformer}), the lack of inter-layer connections led to a rapid decrease in model performance.

In addition to the aforementioned experiments, we conducted an ablation study on the hyperparameter $\lambda$, as denoted in the loss function of the main manuscript. By performing a grid search in the range between 0 and 1, as shown in~\Cref{tab:lambda}, we observed the optimal performance when lambda is set to 0.75. It is noteworthy that in the penultimate row of the table, even when introducing the contrastive learning (CL) loss term with a relatively small weight of 0.25, we still achieved a significant improvement. This evidence highlights the effectiveness of the CL approach in the training process of the \HCMC model.

\begin{minipage}[t]{0.49\textwidth}
    % \centering
    \captionof{table}{Results on MLP training separately.}
    \resizebox{\linewidth}{!}{%
    \begin{tabular}{l|cc|cc}
    \toprule
        & \multicolumn{2}{c|}{in-domain} & \multicolumn{2}{c}{out-domain} \\ 
        & pair. & triplet. & pair. & triplet. \\ \hline
       frame &59.75 & 22.67 & 59.98 & 23.06 \\ 
       shot  &58.40 & 21.58 & 58.97 & 21.23 \\
       scene &58.93 & 20.43 & 58.66 & 20.67 \\  \hline
       frame+shot &51.21 & 17.95 & 51.54 & 18.01 \\
       frame+scene &50.94 & 18.14 & 51.49 & 17.79 \\
       frame+shot+scene &50.10 & 16.54 & 50.54 & 16.68 \\ \bottomrule
    \end{tabular}
    }
    \label{tab:main_mlp}
\end{minipage}
\hfill
\begin{minipage}[t]{0.49\textwidth}
    % \centering
    \captionof{table}{Results on Transformer training separately.}
    \resizebox{\linewidth}{!}{%
    \begin{tabular}{l|cc|cc}
    \toprule
        & \multicolumn{2}{c|}{in-domain} & \multicolumn{2}{c}{out-domain} \\
         & pair. & triplet. & pair. & triplet. \\ \hline
       frame &61.12 &23.01  &63.73 &23.93 \\
       shot &59.43 & 22.44 & 59.54 & 22.25 \\
       scene & 59.52 & 21.30 & 60.12 & 22.06 \\ \hline
       frame+shot  & 51.99 & 18.12 & 52.28 & 18.67 \\ 
       frame+scene  & 51.54 & 17.55 & 51.83 & 18.32 \\ 
       frame+shot+scene  & 50.65 & 17.06 & 51.52 & 17.11 \\ \bottomrule
    \end{tabular}
    }
    \label{tab:main_transformer}
\end{minipage}

% ------------

\begin{minipage}[t]{0.49\textwidth}
    \centering
    \captionof{table}{Ablation study on CL loss weight $\lambda$.}
    % \resizebox{\linewidth}{!}{%
    \begin{tabular}{l|cc|cc}
    \toprule
        \multirow{2}*{$\lambda$}& \multicolumn{2}{c|}{in-domain} & \multicolumn{2}{c}{out-domain} \\
         & pair. & triplet. & pair. & triplet. \\ \hline
       1.00 &   55.40 & 21.58 & 54.97 & 21.23 \\
       0.75 &   55.93 & 22.20 & 55.21 & 21.80 \\
       0.50 &   54.28 & 21.33 & 54.68 & 21.02 \\ 
       0.25  &  53.98 & 19.82 & 53.31 & 20.47 \\ 
       0.00  &  52.63 & 18.79 & 52.74 & 18.06 \\  \bottomrule
    \end{tabular}
    % }
    \label{tab:lambda}
\end{minipage}
\hfill
\begin{minipage}[t]{0.49\textwidth}
\centering
    \captionof{figure}{\textbf{Pairwise score ($\beta=2$ in OS) of reordering results \wrt frame length.}} 
    \label{fig:frame_length}
    \includegraphics[width=\linewidth]{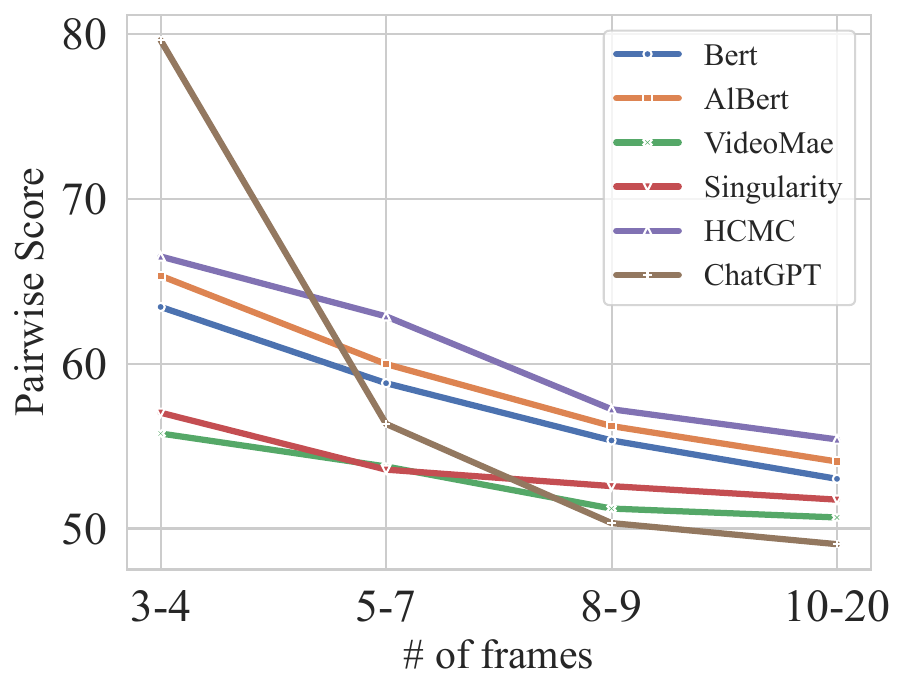}
\end{minipage}

\section{More Numerical Results}\label{sec:more_numerical_results}

\paragraph{Video Lengths}
To evaluate performance on extended videos, we implement a variety of models across a diverse range of frame sequences. As demonstrated in \cref{fig:frame_length}, The \HCMC model, surpassing other models in multi-frame scenarios with over 5 frames, demonstrates that the effective utilization of visual narrative structures can significantly enhance holistic video understanding, which indicates Leveraging visual narrative structures is more important in long-form videos. Additionally, \GPT-3.5-turbo, excelling within the 3-4 frame ordering, displays limitations under regular lengths. We surmise that this is attributable to their insufficient training on tasks involving reordering, which suggests a necessity for further training to enhance their capabilities in long dialogue temporal learning.

\paragraph{Clip Length} We conducted an experiment to compare the impact of different sequence lengths on the performance of our model, as illustrated in the~\Cref{fig:supp_clip_length} below. As the sequence length increases, the overall accuracy of the model demonstrates a declining trend, particularly within the range of 13 to 17 where the decline is more pronounced.
These results suggest that the model's ability to process sequences diminishes as the length of the sequence increases.

\begin{figure}[t!]
    \centering
\begin{subfigure}{.49\linewidth}
    \centering
    \includegraphics[width=1\linewidth]{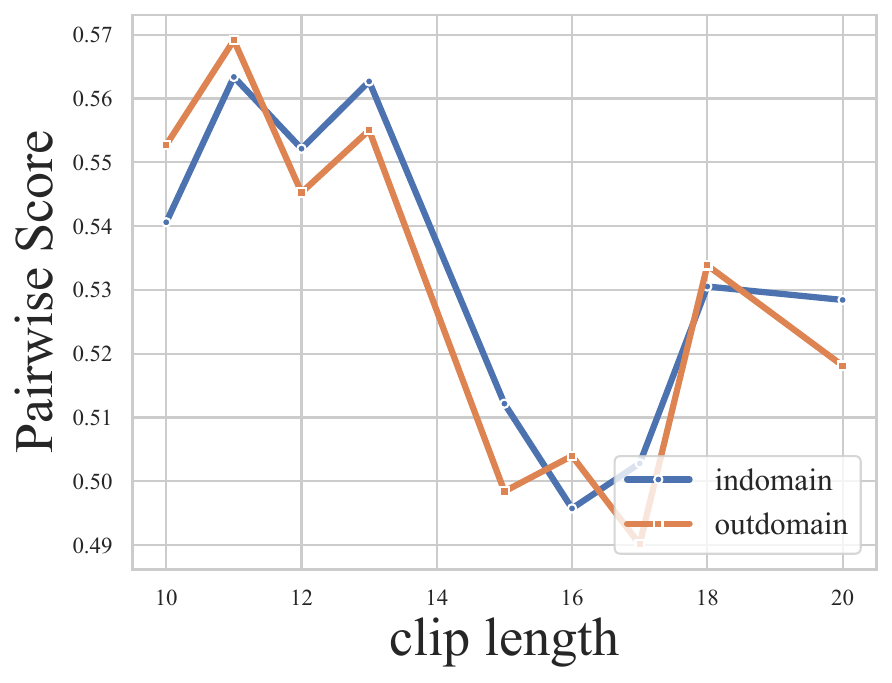}
    % \vspace*{-0mm}
    \caption{Pairwise Score. } 
    \label{fig:len_pair}
\end{subfigure}
\hfill
\begin{subfigure}{.49\linewidth}
    \centering
    \includegraphics[width=1\linewidth]{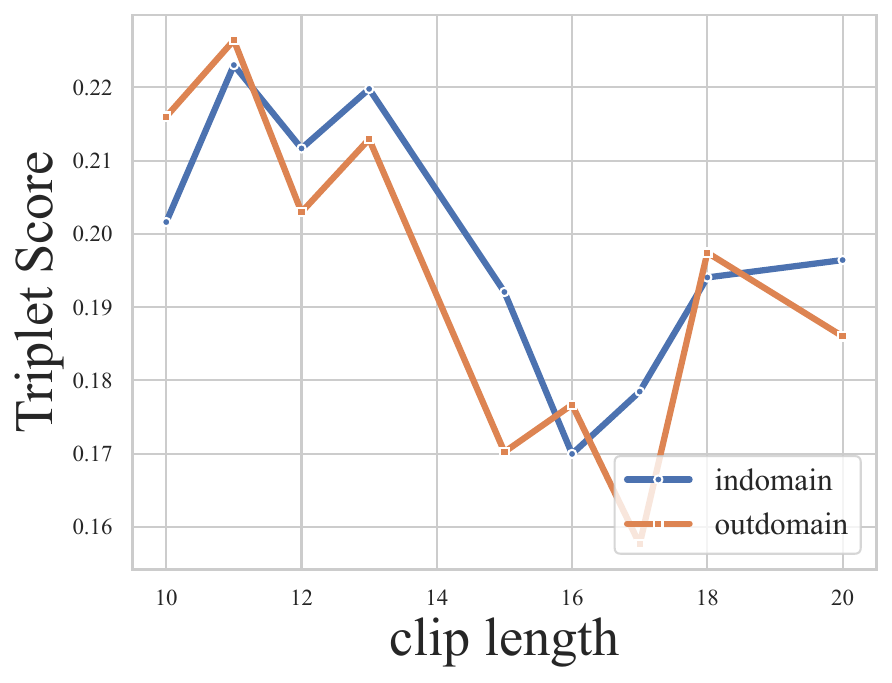}
    % \vspace*{-0mm}
    \caption{Triplet Score. } 
    \label{fig:len_trip}
\end{subfigure}
\caption{Ablated results on clip length.}
\label{fig:supp_clip_length}
\end{figure}

\begin{figure}[t!]
    \centering
\begin{subfigure}{.48\linewidth}
    \centering
    \includegraphics[width=\linewidth]{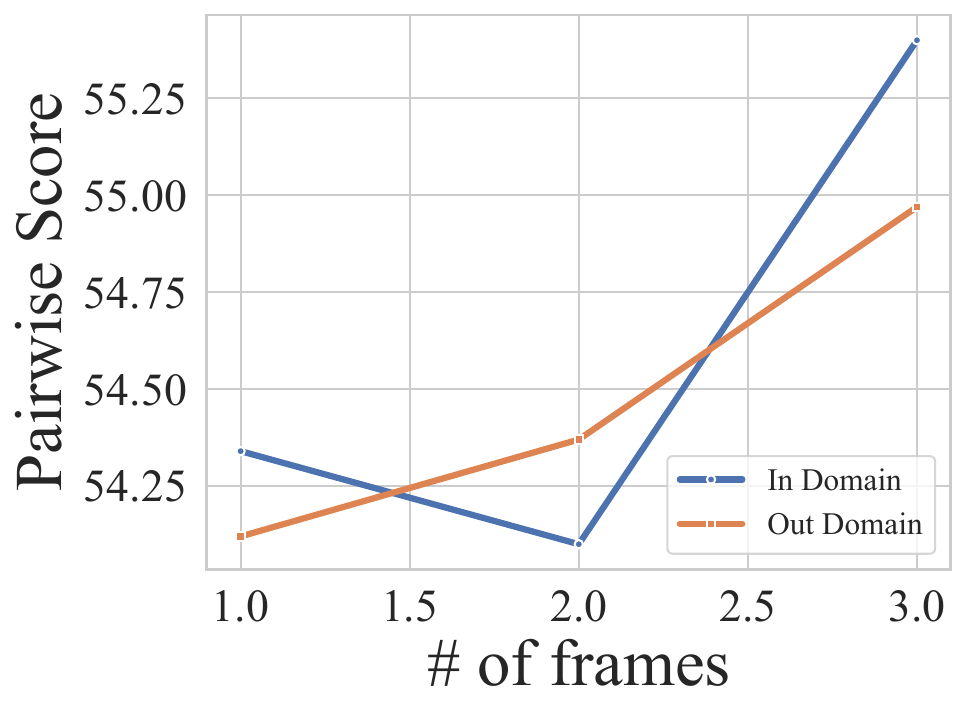}
    % \vspace*{-0mm}
    \caption{Shot Length. } 
    \label{fig:shot-lenght}
\end{subfigure}
\hfill
\begin{subfigure}{.48\linewidth}
    \centering
    \includegraphics[width=\linewidth]{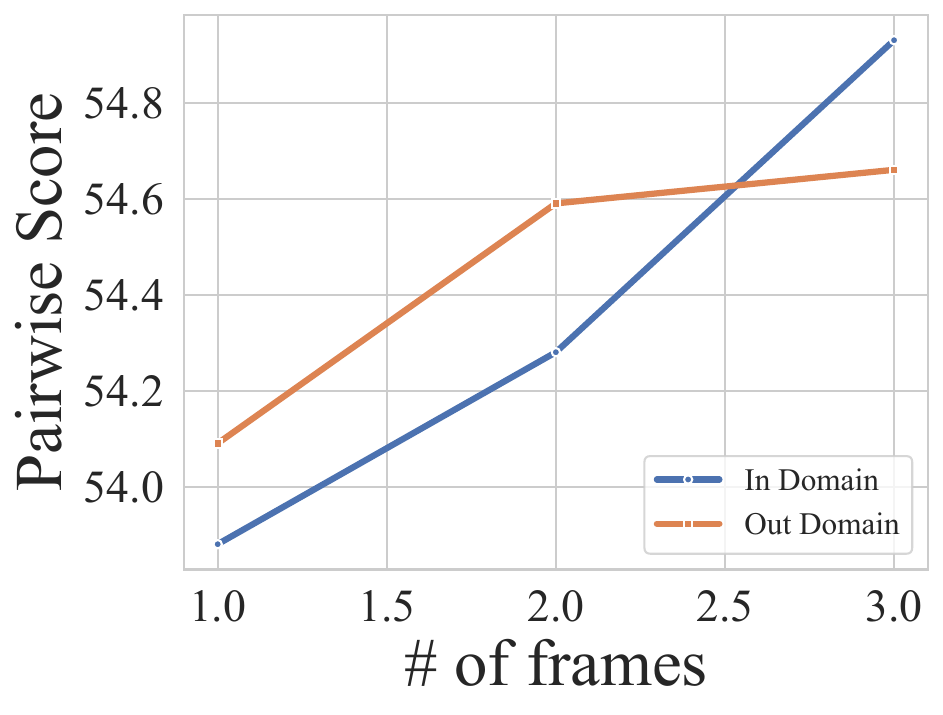}
    % \vspace*{-0mm}
    \caption{Scene Length. } 
    \label{fig:scene-length}
\end{subfigure}
\caption{Ablated results on video sampling lengths.}
\end{figure}

We also investigate the impact of the number of hierarchical structures on experimental results through ablation experiments.
\Cref{fig:scene-length} shows the results of running the frame+shot \HCMC~model with the lengths of frames within shots controlled at 1, 2, and 3 in the entire video clip. 
For the in-domain test dataset, the accuracy slightly decreases when the length increases from 1 to 2, but then rapidly increases to its maximum value when the length increases to 3. For the out-domain test set, the model's performance gradually improves with the increasing complexity of the data.
The results presented in \Cref{fig:shot-lenght} depict the outcome of utilizing the frame+scene \HCMC~model to manipulate the lengths of frames within shots at 1, 2, and 3 in the entire video clip.
For both the in-domain and out-domain test datasets, the accuracy increases with the increasing length of frames in scenes.
From the ablation experiments, it can be concluded that, in general, data with hierarchical structures can be helpful for the model to understand video clips.

\section{Limitation}\label{sec:limitation}

While the proposed framework demonstrates promising performance across diverse experimental settings, certain challenges remain, particularly in handling long sequences of highly shuffled frames. This is partly attributable to the inherent difficulty of modeling global temporal consistency under computational constraints. In addition, the framework is built upon the assumption that frames with similar visual or contextual cues typically belong to the same shot or scene. However, this assumption may be less robust in cinematic videos, where montage editing techniques often involve frequent shot transitions and intentional disruptions of temporal continuity.

Nevertheless, through the introduction of MoviePuzzle and its preliminary results, we aim to provide new perspectives and inspire future research on more generalized and robust video comprehension systems.

\section{Qualitative Image}\label{sec:qualitative_image}
\begin{figure*}[b!]
    \centering

\begin{subfigure}{1\linewidth}
    \centering
        \includegraphics[width=\linewidth]{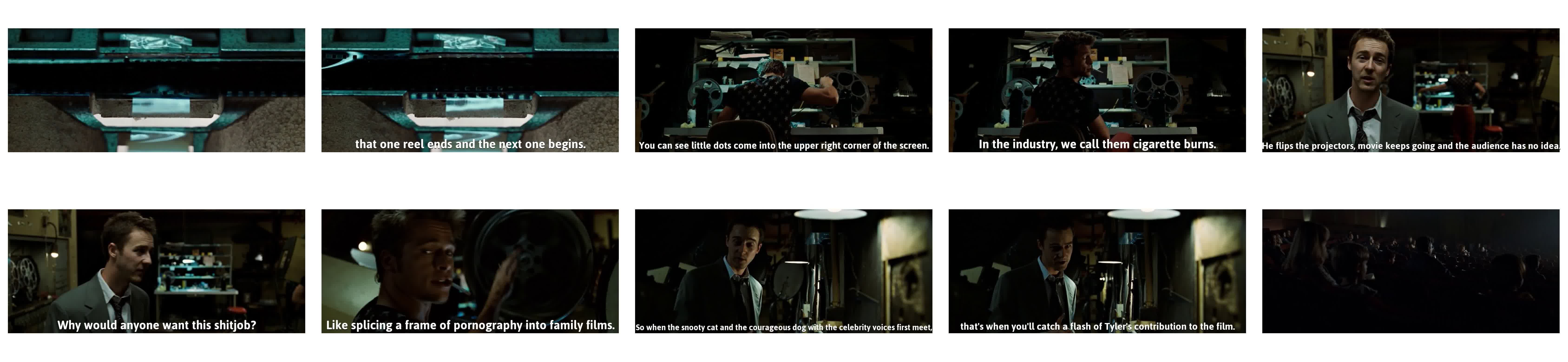}
        \caption{Ground Truth}
        \label{fig:gt5065}
\end{subfigure}

\begin{subfigure}{1\linewidth}
    \centering
        \includegraphics[width=\linewidth]{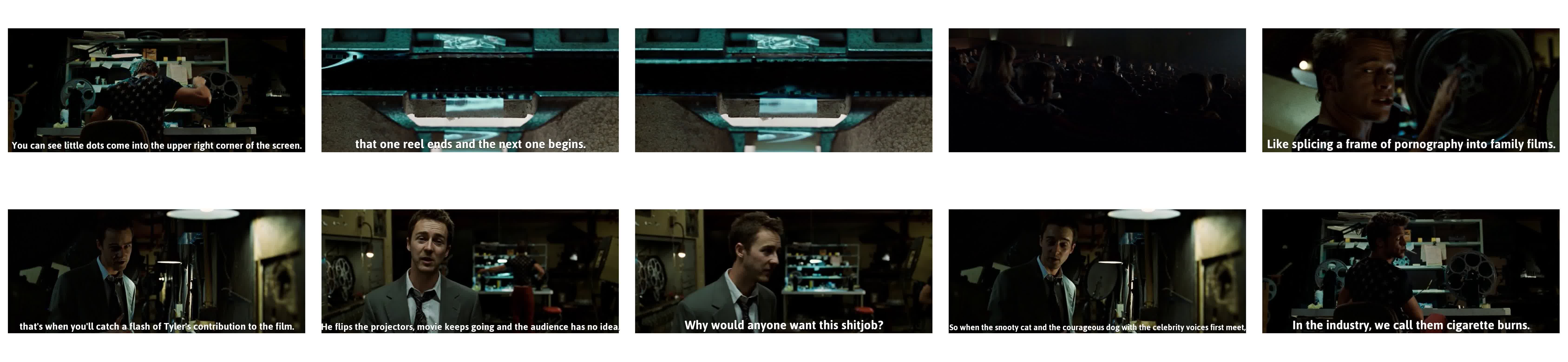}
        \caption{\HCMC}
        \label{fig:ours5065}
\end{subfigure}

\begin{subfigure}{1\linewidth}
    \centering
        \includegraphics[width=\linewidth]{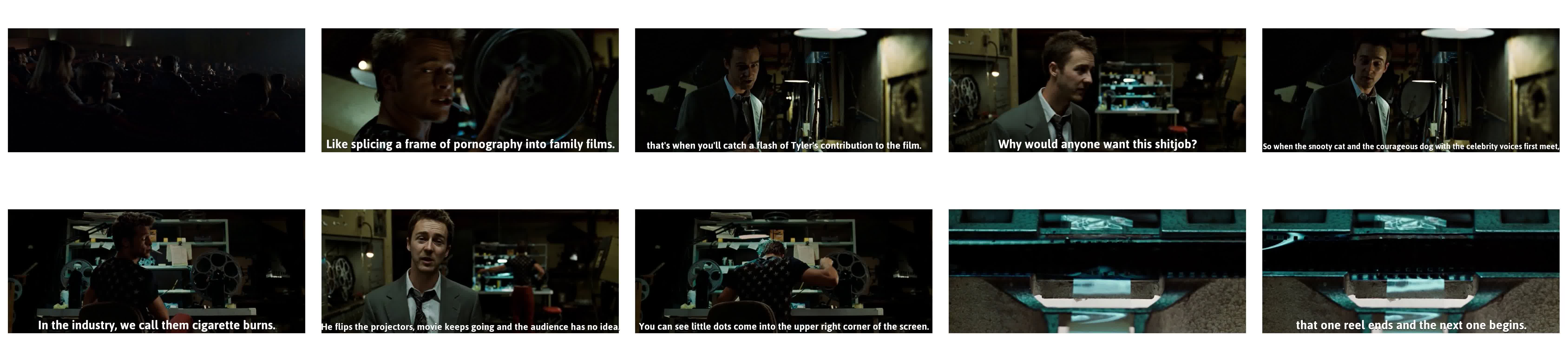}
        \caption{Baseline}
        \label{fig:baseline5065}
\end{subfigure}

%From top to bottom are ground truth, our \HCMC~model and \AlBert~Baseline.
\caption{\textbf{Movie:} \textit{Fight Club}. \textbf{Synopsis:} After one meeting, he confronts her. She argues that she's doing exactly what he does and quips that the groups are 'cheaper than a movie and there's free coffee. Instead of ratting each other out, they agree to split up the week and exchange numbers. Despite his efforts, the narrator's insomnia continues. On a flight back from one of his business trips, the narrator meets Tyler Durden. Tyler offers a unique perspective on emergency procedure manuals in the plane and they strike up a casual conversation. Tyler is a soap salesman, if he's not working nights as a projectionist and slipping bits of porn between reels. The narrator arrives at the baggage claim to discover that his suitcase has been confiscated, most likely due to a mysterious vibration, before he taxis home. However, home, a fifteenth story condominium, has been blasted into the night by what was theorized to be a faulty gas line ignited by a spark on the refrigerator. Having nowhere to go, the narrator finds a business card for Tyler and calls him up. They meet in a parking lot behind a bar where Tyler invites the narrator to ask to come live with him...on one condition: that the narrator hit Tyler as hard as he can. The narrator, though puzzled, complies and they engage in a fist fight before sharing a couple of drinks. The experience is surprisingly euphoric.
}
\label{fig:5065}
\end{figure*}

\begin{figure*}[t!]
    \centering

\begin{subfigure}{1\linewidth}
    \centering
        \includegraphics[width=\linewidth]{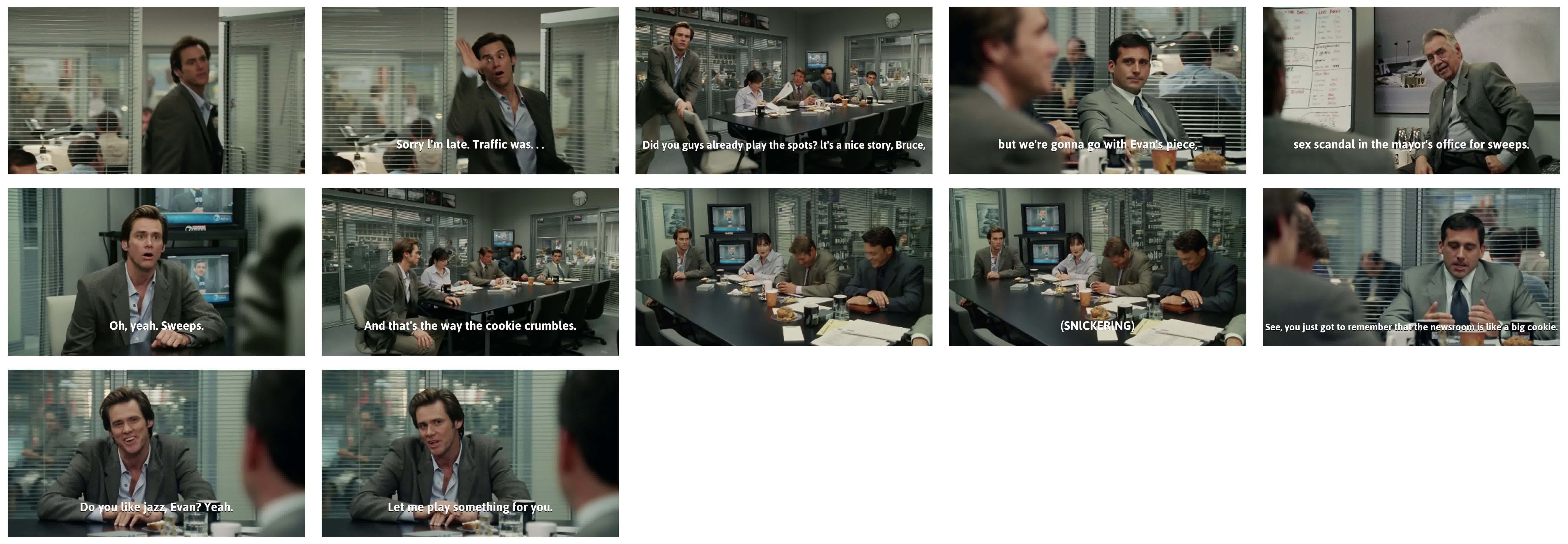}
        \caption{Ground Truth}
        \label{fig:gt6175}
\end{subfigure}

\begin{subfigure}{1\linewidth}
    \centering
        \includegraphics[width=\linewidth]{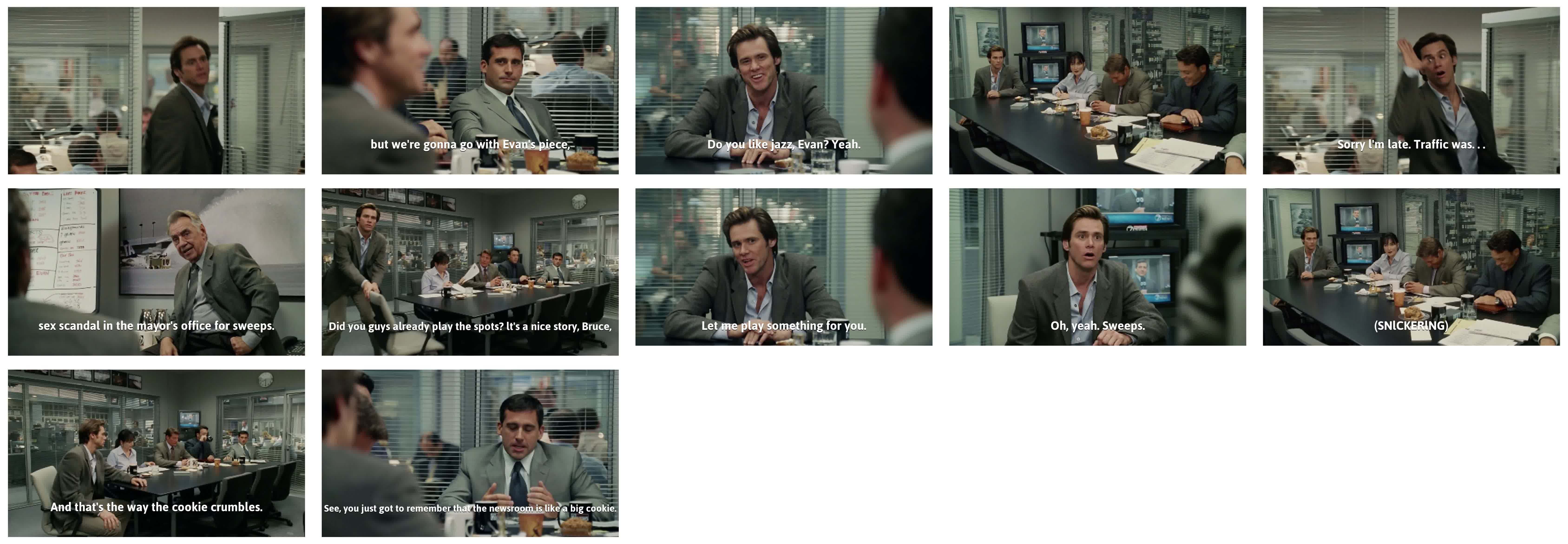}
        \caption{\HCMC}
        \label{fig:ours6175}
\end{subfigure}

\begin{subfigure}{1\linewidth}
    \centering
        \includegraphics[width=\linewidth]{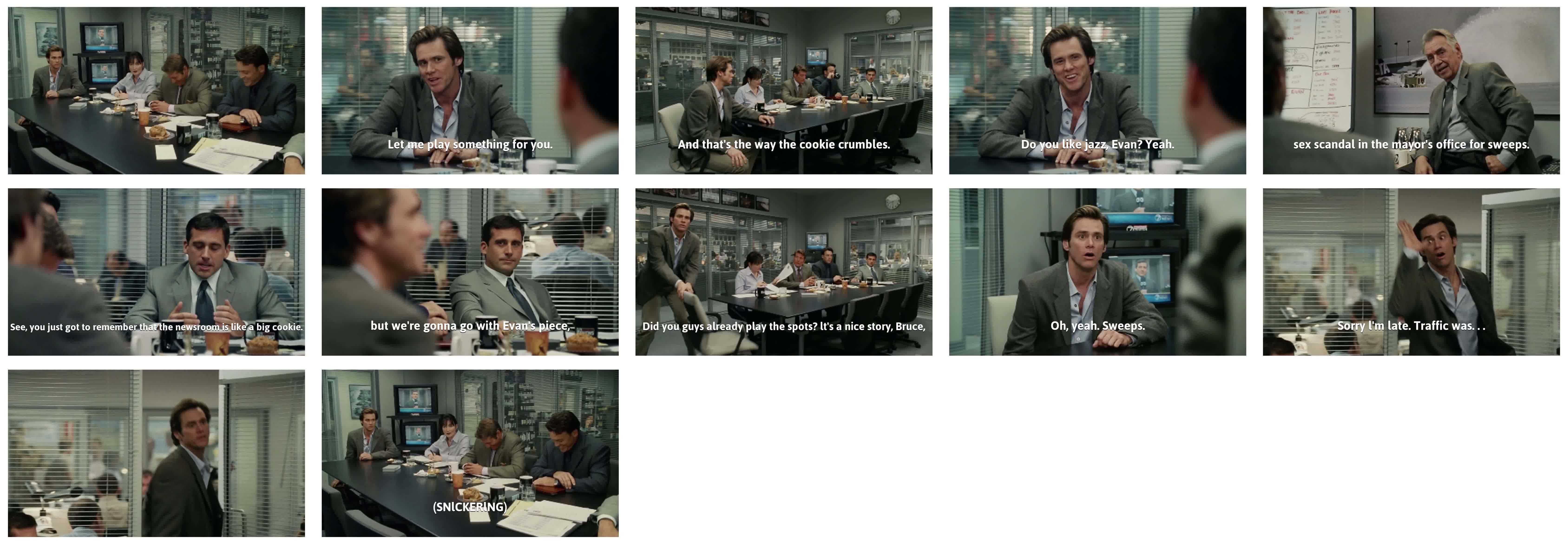}
        \caption{Baseline}
        \label{fig:baseline6175}
\end{subfigure}
    
\caption{\textbf{Movie:} \textit{Bruce Almighty}. 
\textbf{Synopsis:} Racing back to the station, Bruce gets caught in traffic and vents his frustration about the fact that his life is in a go-nowhere rut. Arriving late to an important meeting, fellow staffers--including nemesis Evan Baxter (Steve Carrell)--needle Bruce mercilessly about his clownish coverage at the bakery, further exacerbating his bitterness about being stalled on his career path. In the exchange with Evan, we see that Bruce has a lively, but dark, sense of humor and won't take anything lying down. After the meeting, Bruce begs his boss, Jack Baylor (Philip Baker Hall), to consider him for the open anchor position.}
\label{fig:6175}
\end{figure*}

\begin{figure*}[t!]
    \centering

\begin{subfigure}{1\linewidth}
    \centering
        \includegraphics[width=\linewidth]{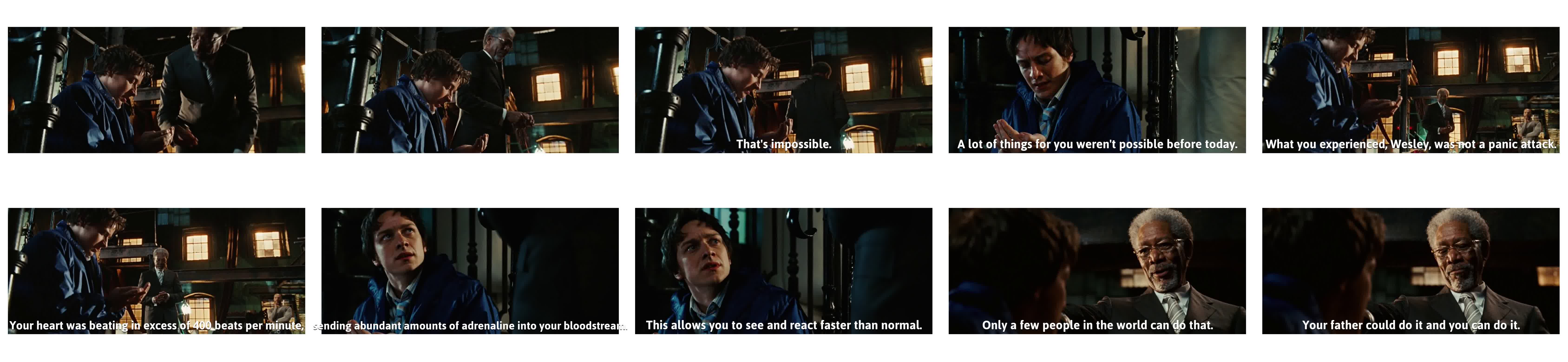}
        \caption{Ground Truth}
        \label{fig:gt7295}
\end{subfigure}

\begin{subfigure}{1\linewidth}
    \centering
        \includegraphics[width=\linewidth]{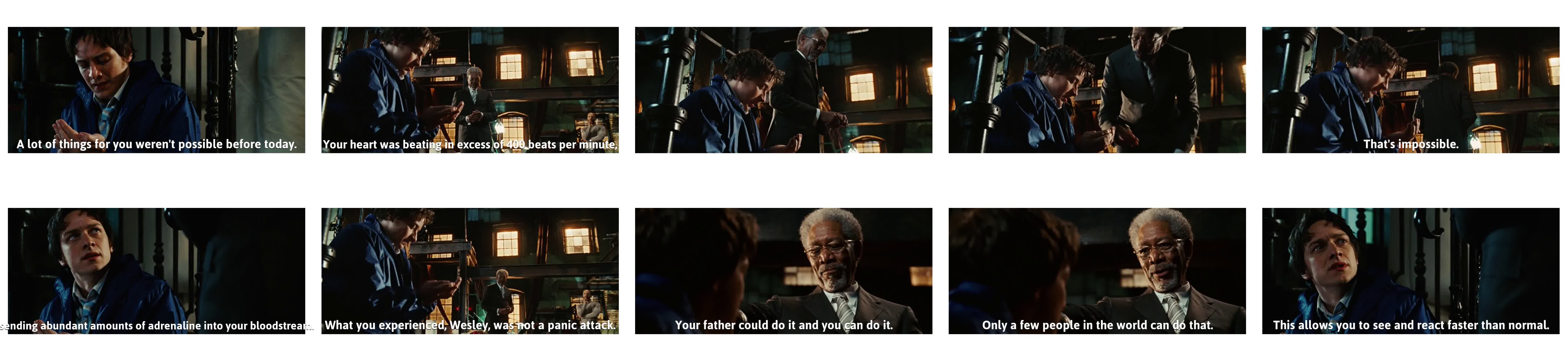}
        \caption{\HCMC}
        \label{fig:ours7295}
\end{subfigure}

\begin{subfigure}{1\linewidth}
    \centering
        \includegraphics[width=\linewidth]{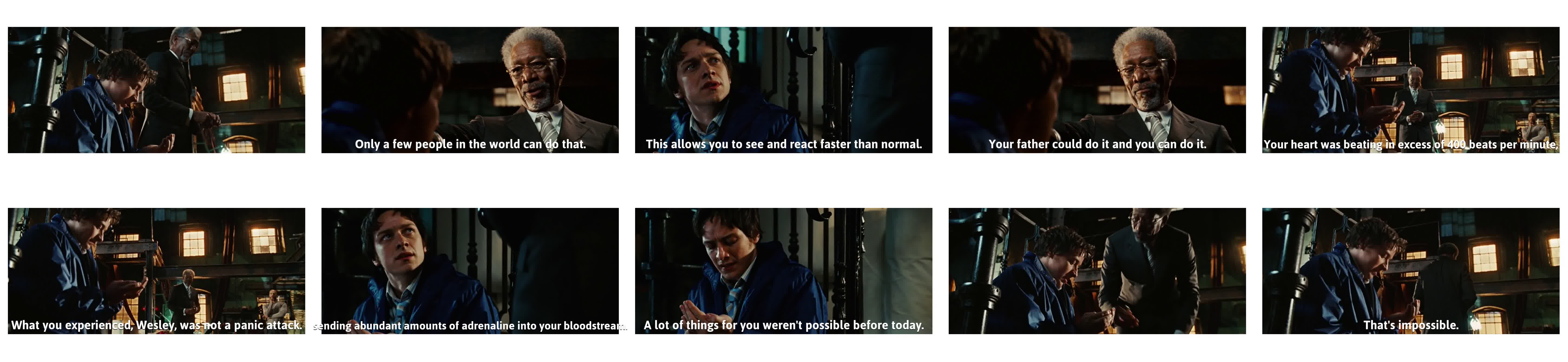}
        \caption{Baseline}
        \label{fig:baseline7295}
\end{subfigure}
    
\caption{\textbf{Movie:} \textit{Wanted}. \textbf{Synopsis:} One night at a pharmacy, Gibson meets a mysterious woman who tells him his father was an elite assassin who had been killed the day before. Gibson replies that his father abandoned him a week after his birth. At that moment, Cross appears, gun in hand. The woman opens fire on Cross. Gibson and the woman escape from the resulting shoot-out and have a wild car chase in the streets of Chicago. The woman brings Gibson to the headquarters of The Fraternity, a thousand-year-old secret society of assassins. The group's leader, Sloan (Morgan Freeman), formally introduces Gibson to Fox (Angelina Jolie), the woman from the night before, and invites him to follow in his father's footsteps as an assassin. Sloan tests Gibson by making him shoot the wings off a fly. When Gibson refuses, a gun is put to his head, triggering a panic attack. Gibson somehow manages to shoot the wings off several flies. Sloan says that he was able to do that because his heart beats 400 times a second when he's stressed. When Sloan asks him whether he want to know how to control it, he runs away in fear. Gibson wakes up the next day hoping everything was a dream, but discovers his father's gun (which he stashes in the toilet tank), and that he has \$3.6 million in his bank account. At work, Gibson tells off his boss, bashes his duplicitous friend with a computer keyboard, and storms out. Gibson then sees pictures of himself and Fox on the front page of several newspapers as wanted fugitives for the pharmacy shooting. Then he notices Fox, who has been waiting outside, and she gives him a ride back to the Fraternity headquarters - an unassuming textile mill.}
\label{fig:7295}
\end{figure*}

\begin{figure*}[t!]
    \centering

\begin{subfigure}{1\linewidth}
    \centering
        \includegraphics[width=\linewidth]{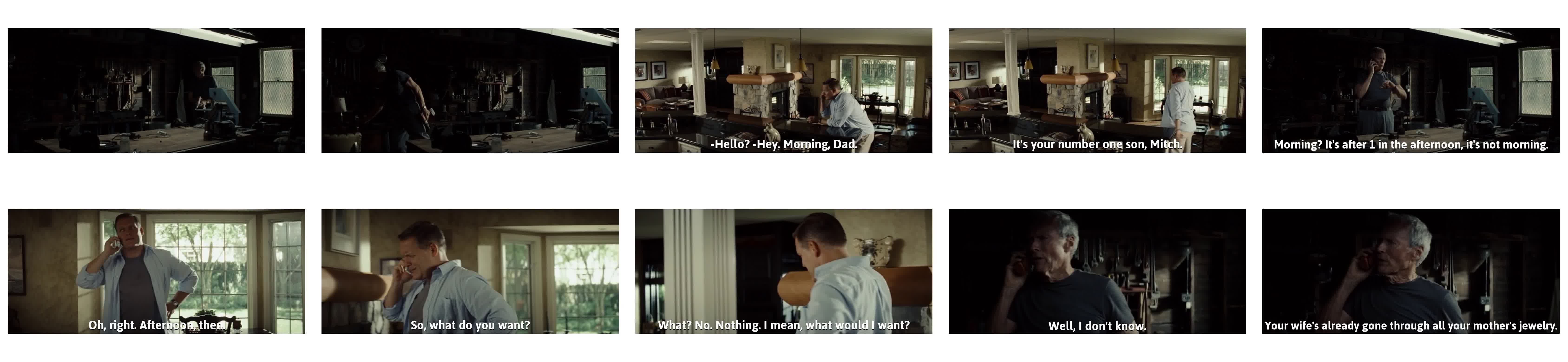}
        \caption{Ground Truth}
        \label{fig:gt8846}
\end{subfigure}

\begin{subfigure}{1\linewidth}
    \centering
        \includegraphics[width=\linewidth]{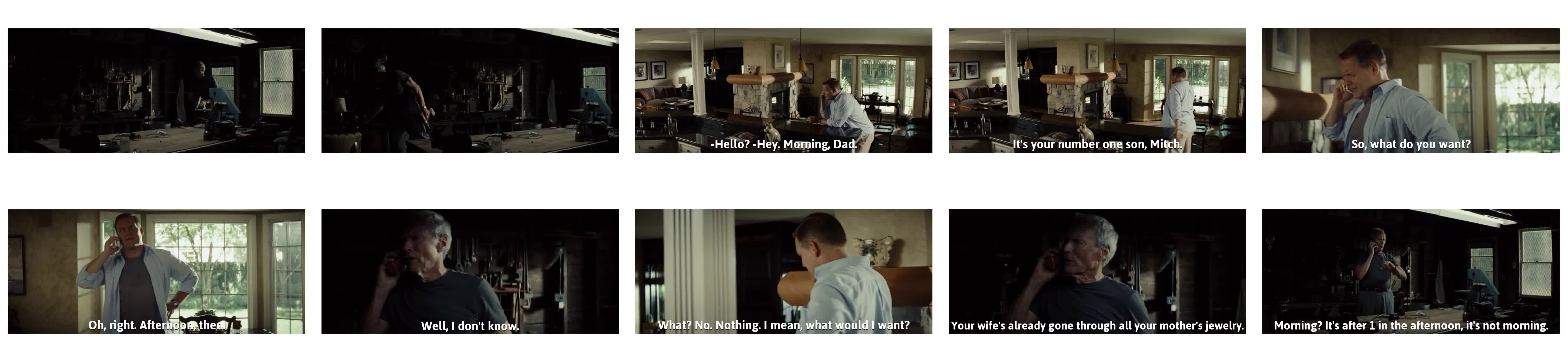}
        \caption{\HCMC}
        \label{fig:ours8846}
\end{subfigure}

\begin{subfigure}{1\linewidth}
    \centering
        \includegraphics[width=\linewidth]{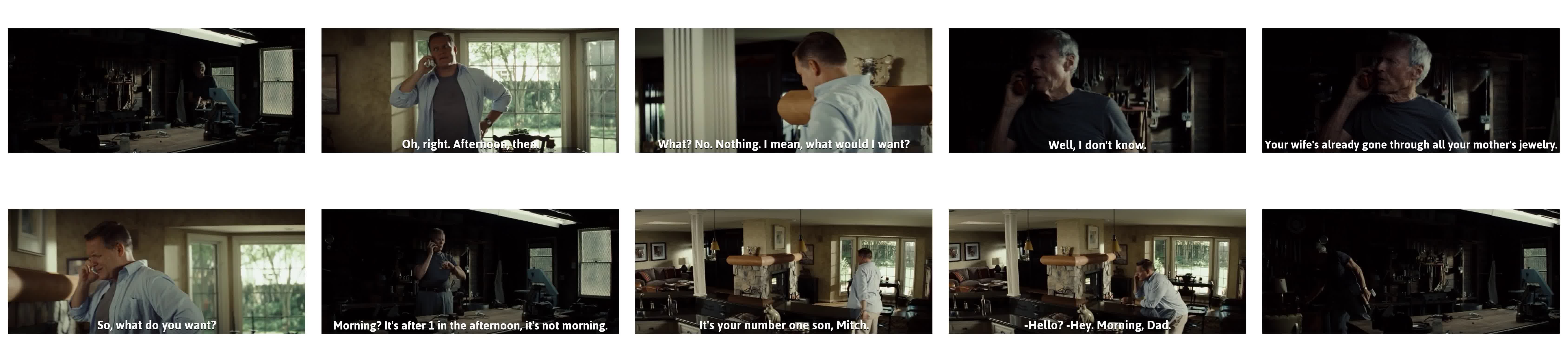}
        \caption{Baseline}
        \label{fig:baseline8846}
\end{subfigure}
    
\caption{\textbf{Movie:} \textit{Gran Torino}. \textbf{Synopsis:}
 % A conversation takes place between a father and his son, Mitch. The son calls his father in the afternoon, initially saying "Morning, Dad" by mistake. The father corrects him and asks what he wants. Mitch seems to be defensive and denies wanting anything. The father then mentions that Mitch's wife has already gone through his late mother's jewelry, implying that the son might have ulterior motives for calling.
 Mitch and his wife, Karen (Geraldine Hughes) go to visit Walt on his birthday, bringing him a cake and a few gifts meant to make certain menial tasks easier. Presentation, and explanation, of these gifts quickly turn into a shamelessly brazen pitch to get Walt to move into a senior's retirement home. Knowing that Mitch and Karen just want to get their hands on his house, Walt growls in anger and throws them out; gifts, cake and all. Mitch and Karen cannot understand Walt's reaction.
}
\label{fig:8846}
\end{figure*}

\begin{figure*}[t!]
    \centering

\begin{subfigure}{1\linewidth}
    \centering
        \includegraphics[width=\linewidth]{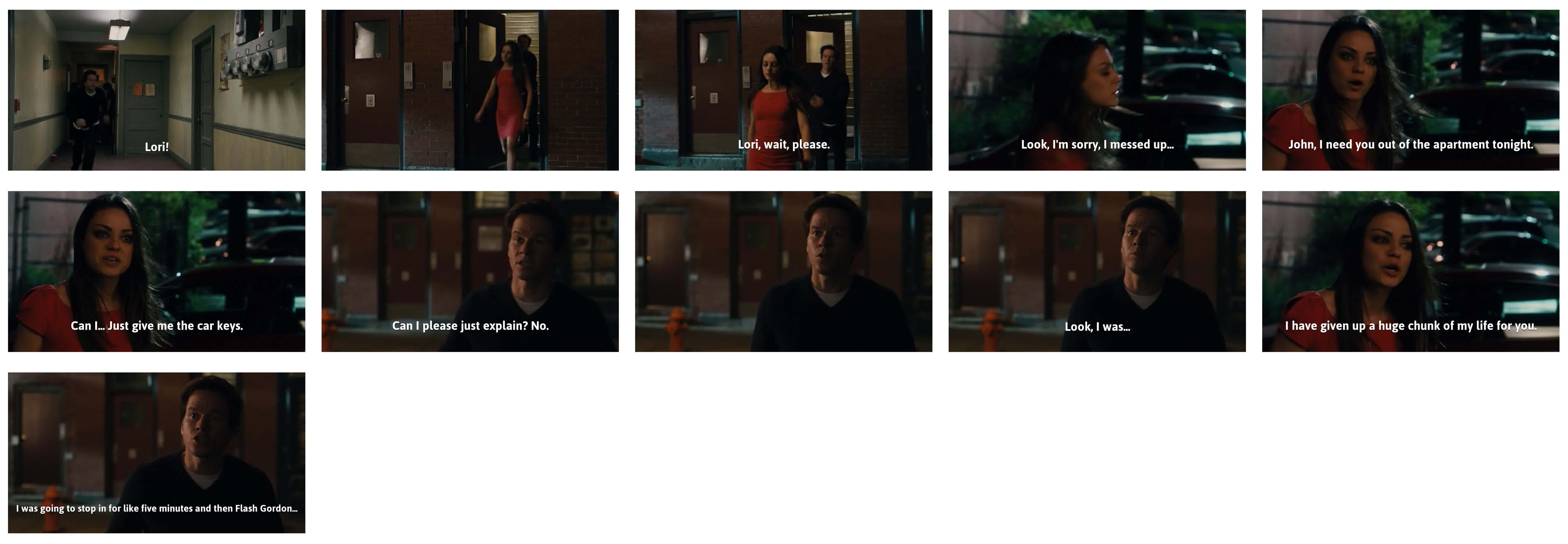}
        \caption{Ground Truth}
        \label{fig:gt9626}
\end{subfigure}

\begin{subfigure}{1\linewidth}
    \centering
        \includegraphics[width=\linewidth]{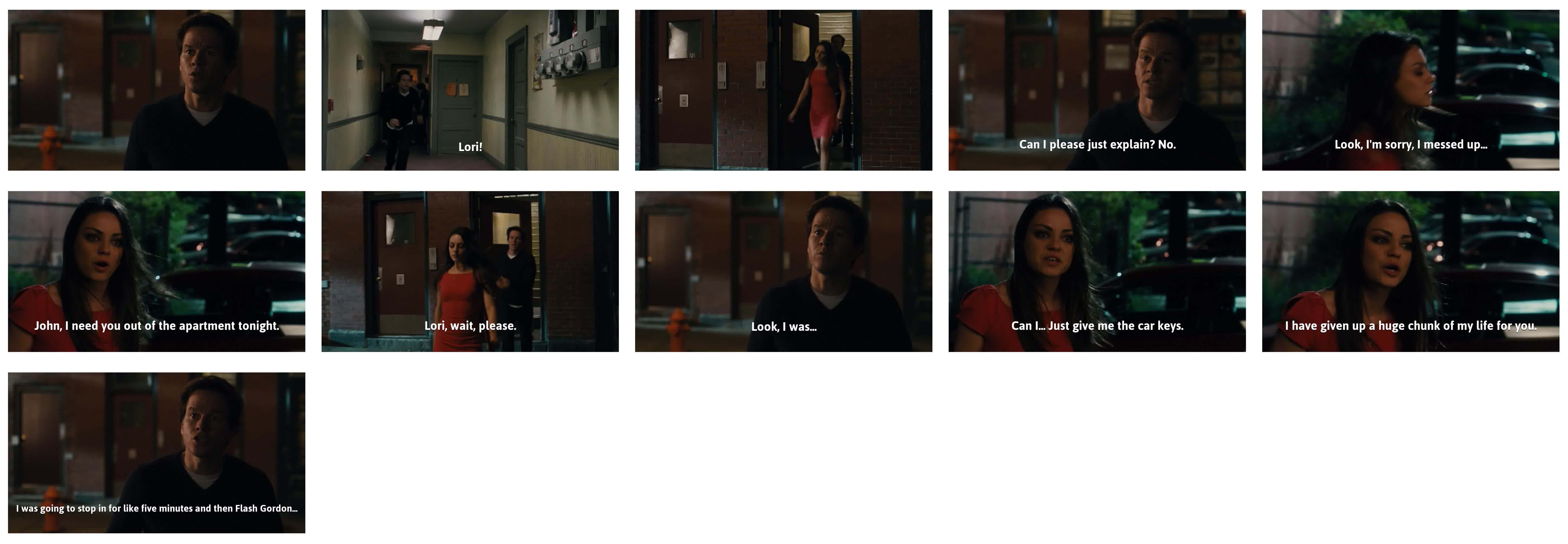}
        \caption{\HCMC}
        \label{fig:ours9626}
\end{subfigure}

\begin{subfigure}{1\linewidth}
    \centering
        \includegraphics[width=\linewidth]{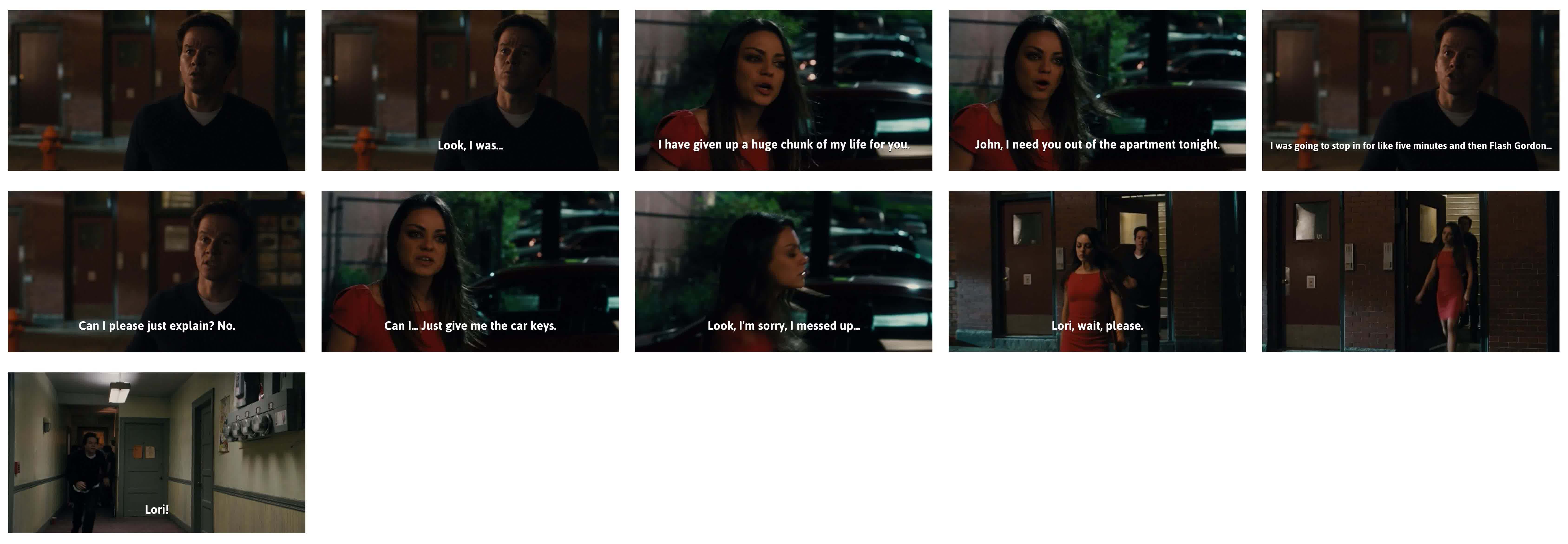}
        \caption{Baseline}
        \label{fig:baseline9626}
\end{subfigure}
    
\caption{\textbf{Movie:} \textit{Ted}. \textbf{Synopsis:} 
John helps Ted secure an apartment and a job at a grocery store, where Ted's reckless behavior earns him both a promotion and a relationship with his colleague Tami-Lynn. Ted's actions, however, strain his friendship with John, particularly when Lori, John's partner, finds out John has been lying about his whereabouts. Concurrently, Donny, a fan of Ted, plots to take him for his son. Things escalate when a party hosted by Lori's boss Rex leads to a wild night at Ted's, causing Lori and John to break up and John to blame Ted for his troubles.}

% John assists Ted in finding an apartment and a job at a grocery store, where Ted's irresponsible behavior surprisingly leads to both a promotion and a relationship with Tami-Lynn, his coworker. This irks Lori, John's partner, who is shocked by Tami-Lynn's temper. Despite this, Ted and John continue to spend a lot of time together, frustrating Lori, especially when she discovers John has been missing work and lying about it. Meanwhile, Donny, a loner who admires Ted, wants to take him for his destructive son. The situation worsens when Lori and John attend a party thrown by her flirtatious boss, Rex. There, Ted entices John to a raucous party at his apartment, featuring their hero Sam J. Jones. John gets swept up in the wild party, leading to a fallout with Lori, who finds him there and ends their relationship. John then blames Ted for his ruined life and distances himself..}
\label{fig:9626}
\end{figure*}

\begin{figure*}[t!]
    \centering

\begin{subfigure}{1\linewidth}
    \centering
        \includegraphics[width=\linewidth]{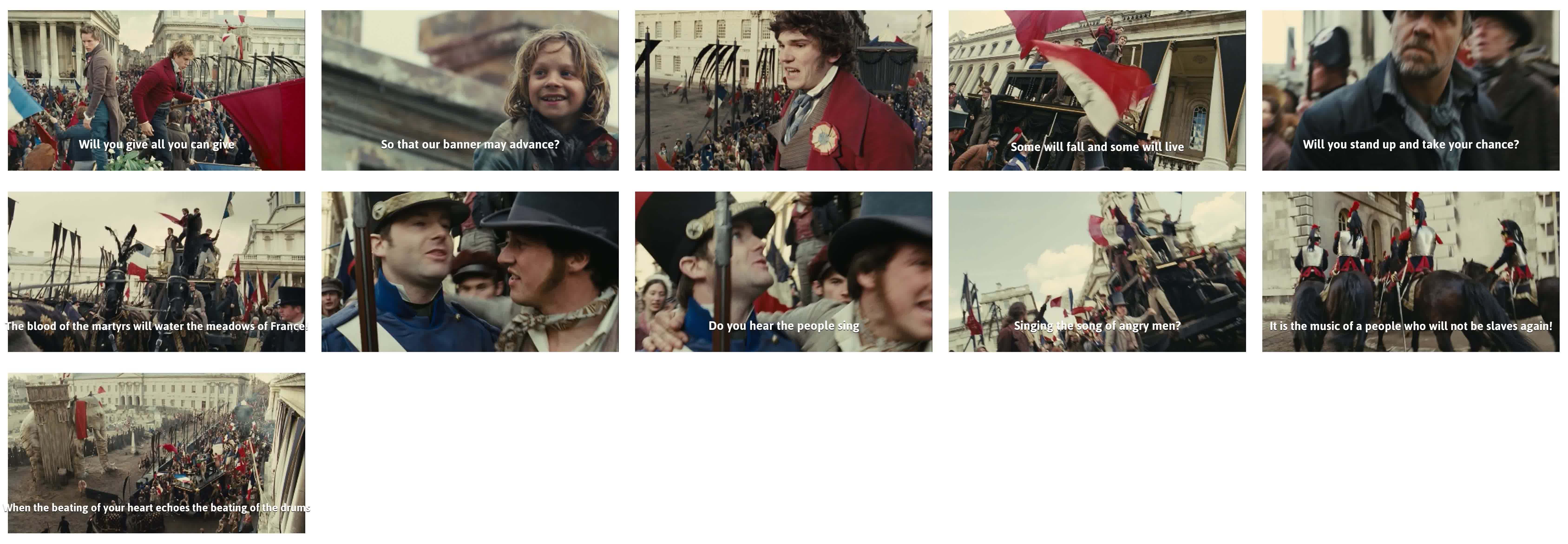}
        \caption{Ground Truth}
        \label{fig:gt9693}
\end{subfigure}

\begin{subfigure}{1\linewidth}
    \centering
        \includegraphics[width=\linewidth]{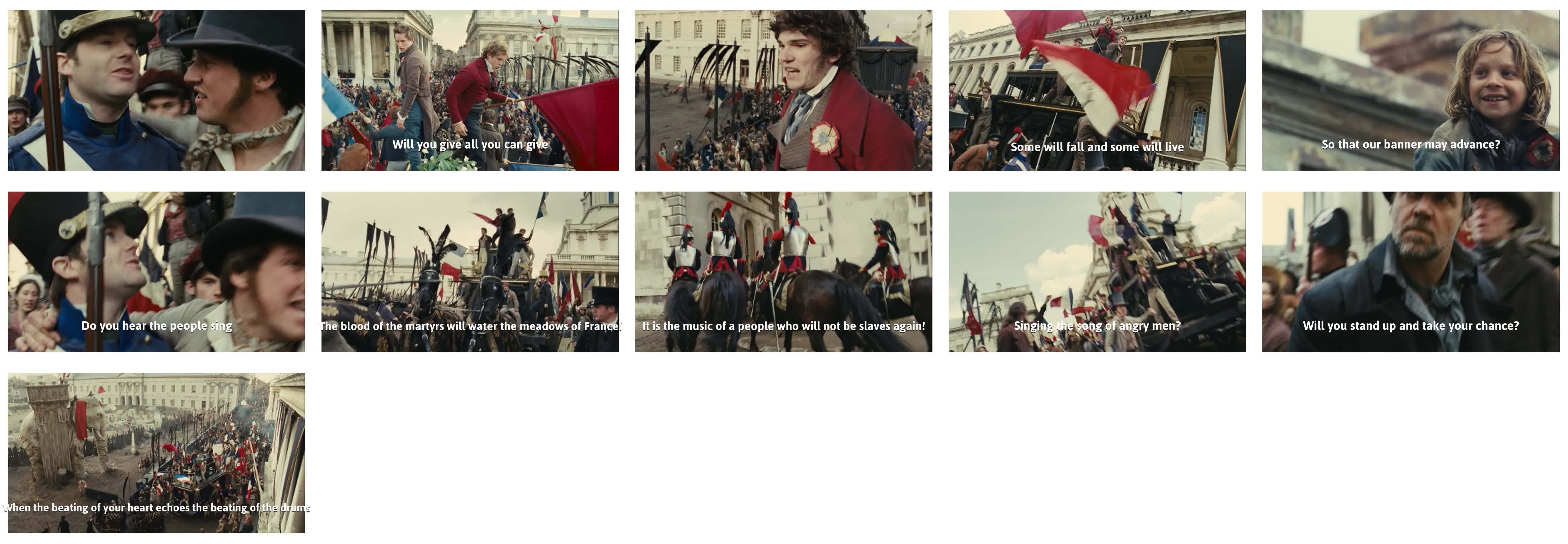}
        \caption{\HCMC}
        \label{fig:ours9693}
\end{subfigure}

\begin{subfigure}{1\linewidth}
    \centering
        \includegraphics[width=\linewidth]{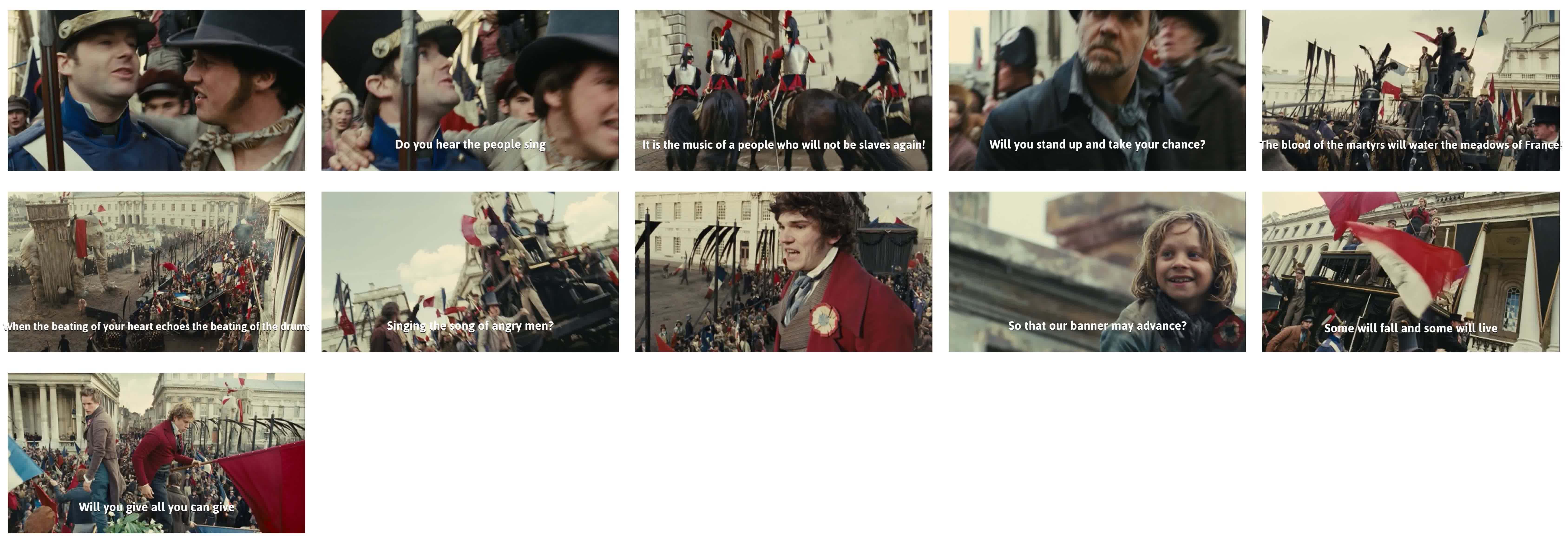}
        \caption{Baseline}
        \label{fig:baseline9693}
\end{subfigure}
    
\caption{\textbf{Movie:} \textit{Les Mis\'erable}. \textbf{Synopsis:} The next day, the students interrupt Lamarque's funeral procession and begin their assault. Javert poses as a rebel in order to spy on them, but is quickly exposed by Gavroche and captured. During the ensuing gunfight, Eponine saves Marius at the cost of her own life, professing her love to him before she dies, which leaves Marius devastated at the loss of his best friend. Valjean, intercepting a letter from Marius to Cosette, goes to the barricade to protect Marius. After saving Enjolras from snipers, he is allowed to execute Javert. However, when the two are alone, Valjean chooses to free Javert instead and fires his gun to fake the execution. Initially disbelieving, Javert wonders at Valjean's generosity.
}
\label{fig:9693}
\end{figure*}

Refer to~\Cref{fig:5065,fig:6175,fig:7295,fig:8846,fig:9626,fig:9693} for qualitative comparisons.~\Cref{fig:5065,fig:6175,fig:7295} come from in domain test data set and~\Cref{fig:8846,fig:9626,fig:9693} come from out domain test dataset. From these examples, it can be observed that our method is able to arrange the same shots closer together within a movie clip compared to the baseline, resulting in an overall better reordering performance.

%%%%%%%%%%%%%%%%%%%%%%%%%%%%%%%%%%%%%%%%%%%%%%%

%%%%%%%%%%%%%%%%%%%%%%%%%%%%%%%%%%%%%%%%%%%%%%%

%%%%%%%%%%%%%%%%%%%%%%%%%%%%%%%%%%%%%%%%%%%%%%%

%%%%%%%%%%%%%%%%%%%%%%%%%%%%%%%%%%%%%%%%%%%%%%%

%%%%%%%%%%%%%%%%%%%%%%%%%%%%%%%%%%%%%%%%%%%%%%%
\newpage
% \newgeometry{
%   textwidth=6.875in,
%   textheight=8.875in,
%   top=0.4in,
%   headheight=0in,
%   headsep=0in,
%   hmargin={1in - 0.304in, 1in - 0.304in},
%   columnsep=0.3125in}
% \restoregeometry

%%%%%%%%%%%%%%%%%%%%%%%%%%%%%%%%%%%%%%%%%%%%%%%
\newpage

\clearpage

\section{Responsibility and License}\label{sec:license}
We bear all responsibility in case of violation of rights and our dataset is under the license of CC BY-NC-SA (Attribution-NonCommercial-ShareAlike).

%%%%%%%%%%%%%%%%%%%%%%%%%%%%%%%%%%%%%%%%%%%%%%%%%%%%%%%%%%%%

\end{document}